\documentclass[nohyperref]{article}

\usepackage[utf8]{inputenc} %

\usepackage{graphicx}

\usepackage[T1]{fontenc}

\IfFileExists{headers/config/showoverfull.config}{
	\overfullrule=1cm
}{
}

\usepackage{etoolbox}

\newbool{includeappendix}
\setbool{includeappendix}{true} %
\IfFileExists{headers/config/noappendix.config}{
	\setbool{includeappendix}{false}
}{}

\newif\ifincludeappendixx
\ifbool{includeappendix}{
	\includeappendixxtrue
}{
	\includeappendixxfalse
}

\usepackage{xr} %
\usepackage{filecontents}

\ifbool{includeappendix}{}{
	\input{appendix-labels-loader}

	\externaldocument{appendix-labels}
}

\newcommand{\eg}{e.g., }
\newcommand{\ie}{i.e., }

\usepackage[usenames, dvipsnames]{color} %
\usepackage{xcolor} %

\definecolor{reference}{HTML}{2CA02C}

\definecolor{my-full-blue}{HTML}{1F77B4}

\definecolor{my-full-orange}{HTML}{FF7F0E}

\definecolor{my-full-green}{HTML}{2CA02C}

\definecolor{my-full-red}{HTML}{d62728}

\definecolor{my-full-purple}{HTML}{9467bd}

\definecolor{my-full-brown}{HTML}{8c564b}

\definecolor{my-full-pink}{HTML}{e377c2}

\definecolor{my-full-gray}{HTML}{7f7f7f}

\definecolor{my-full-olive}{HTML}{bcbd22}

\definecolor{my-full-cyan}{HTML}{17becf}

\colorlet{my-blue}{my-full-blue!30}
\colorlet{my-orange}{my-full-orange!30}
\colorlet{my-green}{my-full-green!30}
\colorlet{my-red}{my-full-red!30}
\colorlet{my-purple}{my-full-purple!30}
\colorlet{my-brown}{my-full-brown!30}
\colorlet{my-pink}{my-full-pink!30}
\colorlet{my-gray}{my-full-gray!30}
\colorlet{my-olive}{my-full-olive!30}
\colorlet{my-cyan}{my-full-cyan!30}

\usepackage{listings}

\usepackage{textcomp}

\usepackage{xcolor}

\usepackage[scaled=0.8]{beramono}

\definecolor{ckeyword}{HTML}{7F0055}
\definecolor{ccomment}{HTML}{3F7F5F}
\definecolor{cstring}{HTML}{2A0099}

\lstdefinestyle{numbers}{
	numbers=left,
	framexleftmargin=20pt,
	numberstyle=\tiny,
	firstnumber=auto,
	numbersep=1em,
	xleftmargin=2em
}

\lstdefinestyle{layout}{
	frame=none,
	captionpos=b,
}

\lstdefinestyle{comment-style}{
	morecomment=[l]//,
	morecomment=[s]{/*}{*/},
	commentstyle={\color{ccomment}\itshape},
}

\lstdefinestyle{string-style}{
	morestring=[b]",%
	morestring=[b]',%
	stringstyle={\color{cstring}},
	showstringspaces=false,%
}

\lstdefinestyle{keyword-style}{
	keywordstyle={\ttfamily\bfseries},
	morekeywords={
		function,
		constructor,
		int,
		bool,
		return,
		returns,
		uint
	},
	morekeywords = [2]{},
	keywordstyle = [2]{\text},
	sensitive=true,
}

\lstdefinestyle{input-encoding}{
	inputencoding=utf8,
	extendedchars=true,
	literate=
	{ℝ}{$\reals$}1%
	{→}{$\rightarrow$}1%
	{α}{$\alpha$}1%
	{β}{$\beta$}1%
	{λ}{$\lambda$}1%
	{θ}{$\theta$}1%
	{ϕ}{$\phi$}1%
}

\lstdefinestyle{escaping}{
	moredelim={**[is][\color{blue}]{\%}{\%}},
	escapechar=|,
	mathescape=true
}

\lstdefinestyle{default-style}{
	basicstyle=\fontencoding{T1}\ttfamily\footnotesize,
	style=numbers,
	style=layout,
	style=comment-style,
	style=string-style,
	style=keyword-style,
	style=input-encoding,
	style=escaping,
	tabsize=2,
	upquote=true
}

\lstdefinelanguage{BASIC}{
	language=C++,
	style=default-style
}[keywords,comments,strings]%

\usepackage{microtype}
\usepackage{graphicx}
\usepackage{subfigure}
\usepackage{booktabs} %
\usepackage{wrapfig}
\usepackage{cuted}
\usepackage{makecell}

\usepackage{array}

\usepackage{hyperref}

\usepackage{multirow}

\usepackage{algorithm}
\usepackage{algorithmic}

\PassOptionsToPackage{square,numbers}{natbib}
\usepackage[final]{neurips2022/neurips_2022}

\usepackage{amsmath,amsfonts,bm}

\def\eqref#1{equation~\ref{#1}}

\def\1{\bm{1}}

\def\ve{{\bm{e}}}

\def\vp{{\bm{p}}}

\def\vt{{\bm{t}}}

\def\vx{{\bm{x}}}

\def\mI{{\bm{I}}}

\def\mW{{\bm{W}}}

\DeclareMathAlphabet{\mathsfit}{\encodingdefault}{\sfdefault}{m}{sl}
\SetMathAlphabet{\mathsfit}{bold}{\encodingdefault}{\sfdefault}{bx}{n}

\DeclareMathOperator*{\argmin}{arg\,min}

\usepackage{amsmath}
\usepackage{amssymb}
\usepackage{mathtools}
\usepackage{amsthm}
\usepackage{dsfont}

\usepackage[capitalize,noabbrev]{cleveref}
\hypersetup{
    colorlinks,
    linkcolor={black},
}

\theoremstyle{plain}

\theoremstyle{definition}

\theoremstyle{remark}

\newcommand{\tool}{LAMP}
\newcommand{\reals}{\ensuremath{\mathds{R}}}

\newcommand{\bertbase}{$\text{BERT}_{\text{BASE}}$}
\newcommand{\bertlarge}{$\text{BERT}_{\text{LARGE}}$}

\newcommand{\tinysix}{$\text{TinyBERT}_{6}$}
\newcommand{\lampcos}{$\text{LAMP}_{\text{Cos}}$}
\newcommand{\lampell}{$\text{LAMP}_{L_2+L_1}$}

\newcommand{\lampltwo}{$\text{LAMP}_{L_2}$}

\newcommand{\lampnoperp}{$\text{LAMP}_{\alpha_{\text{lm}}=0}$}
\newcommand{\lampnoreg}{$\text{LAMP}_{\alpha_{\text{reg}}=0}$}
\newcommand{\lampnoswaps}{$\text{LAMP}_{\text{NoDiscrete}}$}
\newcommand{\lampdiscafter}{$\text{LAMP}_{\text{DiscreteAtEnd}}$}

\newcommand{\cola}{CoLA}
\newcommand{\sst}{SST-2}
\newcommand{\rotten}{RottenTomatoes}

\usepackage[capitalize]{cleveref}

\makeatletter
\newcommand\theHALG@line{\thealgorithm.\arabic{ALG@line}}
\makeatother

\crefformat{section}{\S#2#1#3}

\crefrangeformat{section}{\S#3#1#4\crefrangeconjunction\S#5#2#6}

\crefmultiformat{section}{\S#2#1#3}{\crefpairconjunction\S#2#1#3}{\crefmiddleconjunction\S#2#1#3}{\creflastconjunction\S#2#1#3}

\newcommand{\crefrangeconjunction}{--}

\crefname{listing}{Lst.}{listings}
\crefname{line}{Lin.}{Lin.}
\crefname{appendix}{App.}{App.}
\crefname{equation}{Eq.}{Eqs.}
\crefname{section}{Sec.}{Sections}

\definecolor{refcolor}{RGB}{255,0,0} 

\crefformat{figure}{Fig.~#2{\color{refcolor}#1}#3}
\Crefformat{figure}{Fig.~#2{\color{refcolor}#1}#3}
\crefformat{appendix}{App.~#2{\color{refcolor}#1}#3}
\Crefformat{appendix}{App.~#2{\color{refcolor}#1}#3}
\crefformat{equation}{Eq.~#2{\color{refcolor}#1}#3} %
\Crefformat{equation}{Eq.~#2{\color{refcolor}#1}#3} %
\crefformat{section}{Sec.~#2{\color{refcolor}#1}#3}
\Crefformat{section}{Sec.~#2{\color{refcolor}#1}#3}
\crefformat{table}{Table~#2{\color{refcolor}#1}#3}
\Crefformat{table}{Table~#2{\color{refcolor}#1}#3}

\crefrangeformat{table}{Tables~#3{\color{refcolor}#1}#4 to~#5{\color{refcolor}#2}#6}
\Crefrangeformat{table}{Tables~#3{\color{refcolor}#1}#4 to~#5{\color{refcolor}#2}#6}

\definecolor{blueout}{RGB}{12, 92, 148}
\hypersetup{citecolor=blueout}

\newcommand{\appref}[1]{%
	\ifbool{includeappendix}{\cref{#1}}{the appendix}%
}
\newcommand{\Appref}[1]{%
	\ifbool{includeappendix}{\cref{#1}}{The appendix}%
}

\title{LAMP: Extracting Text from Gradients with Language Model Priors}

\author{%
Mislav Balunovi\'c\thanks{Equal contribution.}, Dimitar I. Dimitrov\footnotemark[1], Nikola Jovanovi\'c, Martin Vechev \\
\texttt{\{mislav.balunovic,dimitar.iliev.dimitrov,} \\
\texttt{nikola.jovanovic,martin.vechev\}@inf.ethz.ch} \\
Department of Computer Science\\ ETH Zurich\\
}

\begin{document} 

\maketitle

\begin{abstract}
	Recent work shows that sensitive user data can be reconstructed from gradient updates, breaking the key privacy promise of federated learning.
While success was demonstrated primarily on image data, these methods do not directly transfer to other domains such as text.
In this work, we propose LAMP, a novel attack tailored to textual data, that successfully reconstructs original text from gradients.
Our attack is based on two key insights: (i) modeling prior text probability with an auxiliary language model, guiding the search towards more natural text, and (ii)
alternating continuous and discrete optimization, which minimizes reconstruction loss on embeddings, while avoiding local minima by applying discrete text transformations.
Our experiments demonstrate that LAMP is significantly more effective than prior work: it reconstructs 5x more bigrams and $23\%$ longer subsequences on average.
Moreover, we are the first to recover inputs from batch sizes larger than 1 for textual models. These findings indicate that gradient updates of models operating on textual data leak more information than previously thought.

\end{abstract}

\section{Introduction} \label{sec:intro}

Federated learning~\cite{fedsgd} (FL) is a widely adopted framework for training machine learning models in a decentralized way. Conceptually, FL aims to enable training of highly accurate models without compromising client data privacy, as the raw data never leaves client machines. However, recent work \cite{phong2017privacy, dlg,idlg} has shown that the server can in fact recover the client data, by applying a reconstruction attack on the gradient updates sent from the client during training. Such attacks typically start from a randomly sampled input and modify it such that its corresponding gradients match the gradient update originally sent by the client. While most works focus on reconstruction attacks in computer vision, there has comparatively been little work in the text domain, despite the fact that some of the most prominent applications of FL involve learning over textual data, \eg next-word prediction on mobile phones~\cite{ramaswamy2019keyboard}. A key component of successful attacks in vision has been the use of image priors such as total variation~\cite{geiping}. These priors guide the reconstruction towards natural images, which are more likely to correspond to client data. However, the use of priors has so far been missing from attacks on text~\cite{dlg, tag}, limiting their ability to reconstruct real client data.

\begin{figure*}[t]
	\begin{center}
		\centerline{\includegraphics[width=1.0\linewidth]{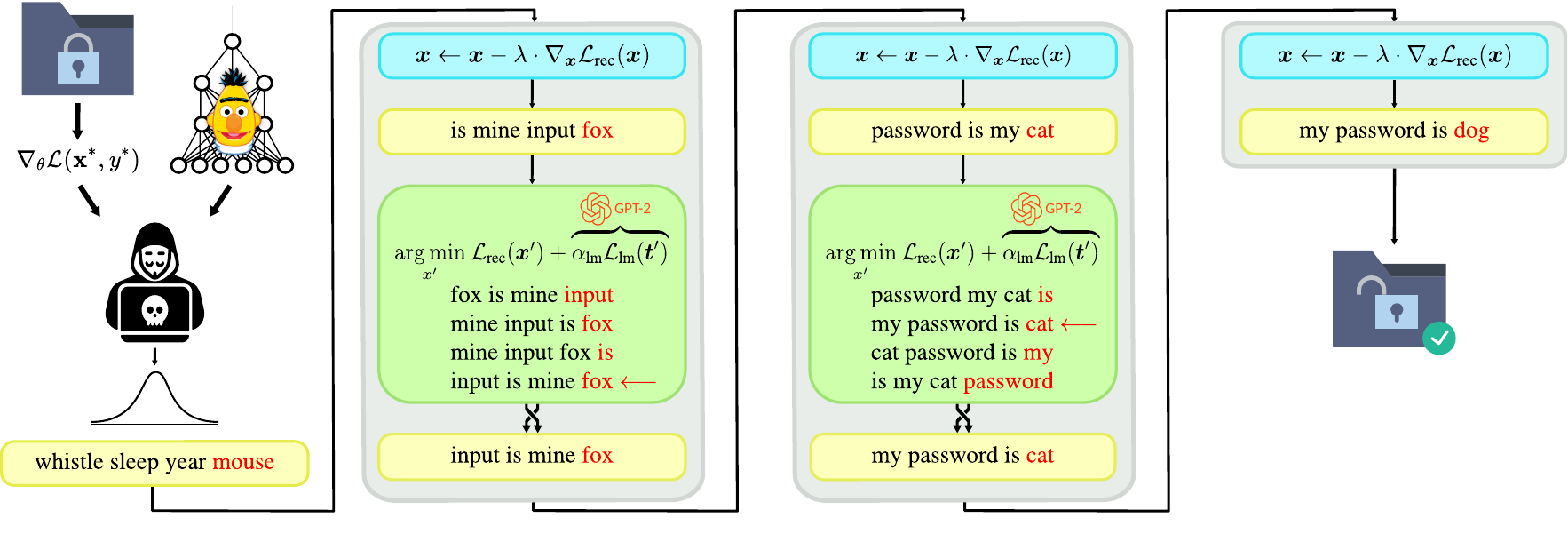}}
	\end{center}
	\vskip -0.31in
	\caption{An overview of \tool{}. We initialize the reconstruction by sampling from a Gaussian distribution, and alternate between continuous and discrete optimization. Continuous optimization minimizes the reconstruction loss with an embedding regularization term. Discrete optimization forms candidates by applying transformations, and chooses the best candidate based on a combination of reconstruction loss and perplexity, as measured by an auxiliary language model (\eg GPT-2).}
	\label{fig:lamp}
\end{figure*}

\paragraph{This work: private text reconstruction with priors}
In this work, we propose LAMP, a new reconstruction attack which leverages language model priors to extract private text from gradients. The overview of our attack is given in~\cref{fig:lamp}. The attacker has access to a snapshot of the transformer network being trained in a federated manner (\eg BERT), and a gradient $\nabla_\theta\mathcal{L}(\vx^*,y^*)$ which the client has computed on that snapshot, using their private data. The attack starts by sampling token embeddings from a Gaussian distribution to create the initial reconstruction. Then, at each step, we improve the reconstruction (shown in yellow) by alternating between continuous (blue) and discrete optimization (green).
The continuous part minimizes the reconstruction loss, which measures how close the gradients of the current reconstruction are to the observed client gradients, together with an embedding regularization term. However, this is insufficient as the gradient descent can get stuck in a local optimum due to its inability to make discrete changes to the reconstruction. We address this issue by introducing a discrete step---namely, we generate a list of candidate sentences using several transformations on the sequence of tokens (\eg moving a token) and select a candidate that minimizes the combined reconstruction loss and perplexity, which measures the likelihood of observing the text in a natural distribution. We use GPT-2~\cite{radford2019gpt2} as an auxiliary language model to measure the perplexity of each candidate (however, our method allows using other models). Our final reconstruction is computed by setting each embedding to its nearest neighbor from the vocabulary.

Key component of our reconstruction attack is the use of a language model prior combined with a search that alternates continuous and discrete optimization steps. Our experimental evaluation demonstrates the effectiveness of this approach---\tool{} is able to extract text from state-of-the-art transformer models on several common datasets, reconstructing up to $5$ times more bigrams than prior work. Moreover, we are the first to perform text reconstruction in more complex settings such as batch sizes larger than 1, fine-tuned models, and defended models. Overall, across all settings we demonstrate that \tool{} is effective in reconstructing large portions of private text.

\paragraph{Main contributions}

Our main contributions are:

\begin{itemize}
\item \tool{}, a novel attack for recovering input text from gradients, which leverages an auxiliary language model to guide the search towards natural text, and a search procedure which alternates continuous and discrete optimization.
\item An implementation of \tool{} and its extensive experimental evaluation, demonstrating that it can reconstruct significantly more private text than prior work. We make our code publicly available at \url{https://github.com/eth-sri/lamp}.
\item The first thorough experimental evaluation of text attacks in more complex settings such as larger batch sizes, fine-tuned models and defended models.
\end{itemize}

\section{Related Work} \label{sec:related}
Federated learning \cite{fedsgd, fedopt} has attracted substantial interest \cite{advances} due to its ability to train deep learning models in a decentralized way, such that individual user data is not shared during training. Instead, individual clients calculate local gradient updates on their private data, and share them with a centralized server, which aggregates them to update the model \cite{fedsgd}. The underlying assumption is that user data cannot be recovered from gradient updates. Recently, several works \cite{phong2017privacy,dlg,idlg,zhu2021rgap} demonstrated that gradients can in fact still leak information, invalidating the fundamental privacy assumption. Moreover, recent work achieved near-perfect image reconstruction from gradients \cite{geiping, nvidia, jeon2021gradient}. Interestingly, prior work showed that an auxiliary model~\cite{jeon2021gradient} or prior information~\cite{bayesian} can significantly improve reconstruction quality. Finally, \citet{arora} noticed that gradient leakage attacks often make strong assumptions, namely that batch normalization statistics and ground truth labels are known. In our work, we do not assume knowledge of batch normalization statistics and as we focus on binary classification tasks, we can simply enumerate all possible labels.

Despite substantial progress on image reconstruction, attacks in other domains remain challenging, as the techniques used for images rely extensively on domain specific knowledge. In the domain of text, in particular, where federated learning is often applied \cite{shejwalkar2021back}, only a handful of works exist \cite{dlg,tag,april}. DLG \cite{dlg} was first to attempt reconstruction from gradients coming from a transformer; TAG \cite{tag} extended DLG by adding an $L_1$ term to the reconstruction loss; finally, unlike TAG and DLG which are optimization-based techniques, APRIL \cite{april} recently demonstrated an exact gradient leakage technique applicable to transformer networks. However, APRIL assumes batch size of $1$ and learnable positional embeddings, which makes it simple to defend against.
Another attack on NLP is given in~\citet{fowl2022decepticons}, but they use stronger assumption that the server can send malicious updates to the clients.
Furthermore, there is a concurrent work~\citep{gupta2022recovering} on reconstructing text from transformers, but it is limited to the case when token embeddings are trained together with the network.

Finally, there have been several works attempting to protect against gradient leakage. Works based on heuristics~\citep{soteria,precode} lack privacy guarantees and have been shown ineffective against stronger attacks~\citep{bayesian}, while those based on differential privacy do train models with formal guarantees~\citep{dpsgd}, but typically hurt the accuracy of the trained models as they require adding noise to the gradients. We remark that we also evaluate \tool{} on defended networks.

\section{Background} \label{sec:background}

In this section, we introduce the background necessary to understand our work.

\subsection{Federated Learning}
In federated learning, $C$ clients aim to jointly optimize a neural network $f$ with parameters $\bm{\theta}$ on their private data. At iteration $k$, the parameters $\bm{\theta}^k$ are sent to all clients, where each client $c$ executes a gradient update $\nabla_{\bm{\theta}^k}\mathcal{L}(\vx^*_c,y^*_c)$ on a sample $(\vx^*_c,y^*_c)$ from their dataset $(\mathcal{X}_c, \mathcal{Y}_c)$. The updates are sent back to the server and aggregated. While in \cref{sec:eval} we experiment with both FedSGD \cite{fedsgd} and FedAvg \cite{konevcny2016federated} client updates, throughout the text we assume that clients use FedSGD updates:
\begin{equation*}
	\bm{\theta}^{k+1} = \bm{\theta}^k - \frac{\lambda}{C}\sum_{c=1}^C \nabla_{\bm{\theta}^k}\mathcal{L}(\vx^*_c,y^*_c).
\end{equation*}
\paragraph{Gradient leakage attacks} A gradient leakage attack is an attack executed by the server (or a party which compromised it) that tries to obtain the private data $(\vx^*_c,y^*_c)$ of a client using the gradient updates $\nabla_{\bm{\theta}^k}\mathcal{L}(\vx^*_c,y^*_c)$ sent to the server. Gradient leakage attacks usually assume honest-but-curious servers which are not allowed to modify the federated training protocol outlined above. A common approach, adopted by \citet{dlg,idlg,tag} as well as our work, is to obtain the private data by solving the optimization problem:
\begin{equation*}
	\argmin_{(\vx_c,y_c)}\: \delta \left( \nabla_{\bm{\theta}^k}\mathcal{L} \left( \vx^*_c,y^*_c \right),\nabla_{\bm{\theta}^k}\mathcal{L} \left( \vx_c,y_c \right) \right),
\end{equation*}
where $\delta$ is some distance measure and $(\vx_c,y_c)$ denotes dummy data optimized using gradient descent to have similar gradients $\nabla_{\bm{\theta}^k}\mathcal{L}(\vx_c,y_c)$ to true data $(\vx^*_c,y^*_c)$. Common choices for $\delta$ are $L_2$~\cite{dlg}, $L_1$~\cite{tag} and cosine distances~\cite{geiping}. When the true label $y^*_c$ is known, the problem reduces to
\begin{equation*}
 	\argmin_{\vx_c}\: \delta \left( \nabla_{\bm{\theta}^k}\mathcal{L} \left( \vx^*_c,y^*_c \right),\nabla_{\bm{\theta}^k}\mathcal{L} \left( \vx_c,y^*_c \right) \right),
\end{equation*}
which was shown \cite{idlg,arora} to be simpler to solve with gradient descent approaches.
\subsection{Transformer Networks} 
In this paper, we focus on the problem of gradient leakage of text on transformers~\citep{vaswani2017attention}. 
Given some input text, the first step is to tokenize it into tokens from some fixed vocabulary of size $V$.
Each token is then converted to a $1$-hot vector denoted $\vt_1,\vt_2,\hdots, \vt_n\in\reals^V$, where $n$ is the number of tokens in the text. The tokens are then converted to embedding vectors $\vx_1,\vx_2,\hdots, \vx_n\in\reals^d$, where $d$ is a chosen embedding size, by multiplying by a trained embedding matrix $\mW_{\text{embed}}\in \reals^{V\times d}$ \cite{bert}. The rows of $\mW_{\text{embed}}$ represent the embeddings of the tokens, and we denote them as $\ve_1,\ve_2,\hdots, \ve_V\in\reals^d$. In addition to tokens, their positions in the sequence are also encoded using the positional embedding matrix $\mW_{\text{pos}}\in\reals^{P\times d}$, where $P$ is the longest allowed token sequence. The resulting positional embeddings are denoted $\vp_1,\vp_2,\hdots, \vp_n$. For notational simplicity, we denote $\ve=\ve_1,\ve_2,\hdots, \ve_V$, $\vx=\vx_1,\vx_2,\hdots, \vx_n$ and $\vp=\vp_1,\vp_2,\hdots, \vp_n$.
We use the token-wise sum of the embeddings $\vx$ and $\vp$ as an input to a sequence of self-attention layers \cite{vaswani2017attention}. The final classification output is given by the first output of the last self-attention layer that undergoes a final linear layer, followed by a $\tanh$.
\subsection{Calculating Perplexity on Pretrained Models}
In this work, we rely on large pretrained language models, such as GPT-2~\citep{radford2019gpt2}, to assess the quality of the text produced by the continuous part of our optimization. Such models are typically trained to calculate the probability $P(\vt_n\;|\;\vt_1,\vt_2,...,\vt_{n-1})$ of inserting a token $\vt_n$ from the vocabulary of tokens to the end of the sequence of tokens $\vt_1,\vt_2,...,\vt_{n-1}$. Therefore, such models can be leveraged to calculate the likelihood of a sequence of tokens $\vt_1,\vt_2,...,\vt_{n}$, as follows:
\begin{equation*}
 P(\vt_1,\vt_2,...,\vt_{n}) = \prod_{l=0}^{n-1} P( \vt_{l+1}\;|\; \vt_1,\vt_2,...,\vt_{l}).
\end{equation*}
One can use the likelihood, or the closely-related negative log-likelihood, as a measure of the quality of a produced sequence. However, the likelihood depends on the length of the sequence, as probability decreases with length. To this end, we use the perplexity measure \citep{jelinek1977perplexity}, defined as:
\begin{equation*}
	\mathcal{L}_{\text{lm}}(\vt_1,\vt_2,...,\vt_{n}) = - \frac{1}{n} \sum_{l=0}^{n-1} \log  P( \vt_{l+1}\;|\; \vt_1,\vt_2,...,\vt_{l}).
\end{equation*}

In the discrete part of our optimization, we rely on this measure to assess the quality of reconstructed sequences produced by the continuous part.

\section{Extracting Text with LAMP} \label{sec:method}
In this section we describe the details of our attack which alternates between continuous optimization using gradient descent, presented in~\cref{sec:contopt}, and discrete optimization using language models to guide the search towards more natural text reconstruction, presented in~\cref{sec:discopt}.

\subsection{Notation}
We denote the attacked neural network and its parameters with $f$ and $\theta$, respectively. Further, we denote the client token sequence and its label as $(\vt^*, y^*)$, and our reconstructions as $(\vt, y)$. For each token $\vt^*_i$ in $\vt^*$ and $\vt_i$ in $\vt$, we denote their embeddings with $\vx^*_i\in \reals^d$ and $\vx_i\in \reals^d$, respectively.  Moreover, for each token in our vocabulary, we denote the embedding with $\ve_i\in\reals^d$. We collect the individual embeddings $\vx_i$, $\vx^*_i$, and $\ve_i$ into the matrices $\vx\in \reals^{d\times n}$, $\vx^*\in \reals^{d\times n}$ and $\ve\in \reals^{d\times V}$, where $n$ is the number of tokens in $\vt^*$ and $V$ is the size of the vocabulary.

\subsection{Continuous Optimization}
\label{sec:contopt}

We now describe the continuous part of our attack (blue in~\cref{fig:lamp}). Throughout the paper, we assume knowledge of the ground truth label $y^*$ of the client token sequence we aim to reconstruct, meaning \mbox{$y = y^*$}. This assumption is not a significant restriction as we mainly focus on binary classification, with batch sizes such that trying all possible combinations of labels is feasible. Moreover, prior work \citep{geng2021towards, nvidia} has demonstrated that labels can easily be recovered for basic network architectures, which can be adapted for transformers in future work. We initialize our reconstruction candidate by sampling embeddings from a Gaussian and pick the one with the smallest reconstruction loss.

\paragraph{Reconstruction loss}
A key component of our attack is a loss measuring how close the reconstructed gradient is to the true gradient. Assuming an $l$-layer network, where $\bm{\theta}_i$ denotes the parameters of layer $i$, an option is to use the combination of $L_2$ and $L_1$ loss proposed by~\citet{tag},
\begin{align*}
  \mathcal{L}_{\text{tag}}(\vx) = \sum^l_{i=1} || \nabla_{\bm{\theta}_i} f(\vx^*, y^*) - \nabla_{\bm{\theta}_i} f(\vx, y) ||_2 +
  \alpha_{\text{tag}} || \nabla_{\bm{\theta}_i} f(\vx^*, y^*) - \nabla_{\bm{\theta}_i} f(\vx, y) ||_1.
\end{align*}
where $\alpha_{\text{tag}}$ is a hyperparameter.
Another option is to use the cosine reconstruction loss proposed by~\citet{geiping} in the image domain:
\begin{equation*}
	\mathcal{L}_{\text{cos}}(\vx) = 1 - \frac{1}{l} \sum^l_{i=1} \frac{ \nabla_{\bm{\theta}_i} f(\vx^*, y^*) \cdot \nabla_{\bm{\theta}_i} f(\vx, y) } { \|\nabla_{\bm{\theta}_i} f(\vx^*, y^*)\|_2 \|\nabla_{\bm{\theta}_i} f(\vx, y)\|_2}.
\end{equation*}
Naturally, LAMP can also be instantiated using any other loss.
Interestingly, we find that there is no loss that is universally better, and the effectiveness is dataset dependent.
Intuitively, $L_1$ loss is less sensitive to outliers, while cosine loss is independent of the gradient norm, so it works well for small gradients.
Thus, we set the gradient loss $\mathcal{L}_{\text{grad}}$ to either $\mathcal{L}_{\text{tag}}$ or $\mathcal{L}_{\text{cos}}$, depending on the setting.

\paragraph{Embedding regularization}
In the process of optimizing the reconstruction loss, we observe the resulting embedding vectors $\vx_i$ often steadily grow in length. We believe this behavior is due to the self-attention layers in transformer networks that rely predominantly on dot product operations. As a result, the optimization process focuses on optimizing the direction of individual embeddings $\vx_i$, disregarding their length. To address this, we propose an embedding length regularization term:
\begin{equation*}
	\mathcal{L}_{\text{reg}}(\vx) = \left( \frac{1}{n}\sum^n_{i=1}\|\vx_i\|_2 - \frac{1}{V}\sum^V_{j=1}\|\ve_j\|_2 \right)^2.
\end{equation*}
The regularizer forces the mean length of the embeddings of the reconstructed sequence to be close to the mean length of the embeddings in the vocabulary. The final gradient reconstruction error optimized in \tool{}  is given by:
\begin{equation*}
	\mathcal{L}_{\text{rec}}(\vx) = \mathcal{L}_{\text{grad}}(\vx) + \alpha_{\text{reg}}\mathcal{L}_{\text{reg}}(\vx),
\end{equation*}
where $\alpha_{\text{reg}}$ is a regularization weighting factor.

\paragraph{Optimization}

We summarize how described components work together in the setting of continuous optimization. To reconstruct the token sequence $\vt^*$, we first randomly initialize a sequence of dummy token embeddings $\vx=\vx_0\vx_1\hdots \vx_n$, with $\vx_i\in \reals^d$. Following prior work on text reconstruction from gradients \cite{tag,dlg}, we apply gradient descent on $\vx$ to minimize the reconstruction loss $\mathcal{L}_{\text{rec}}(\vx)$. To this end, a second-order derivative needs to be computed, as $\mathcal{L}_{\text{rec}}(\vx)$ depends on the network gradient at $\vx$. Similar to prior work \cite{tag,dlg}, we achieve this using automatic differentiation in Pytorch~\cite{NEURIPS2019_9015}.  

\subsection{Discrete Optimization}
\label{sec:discopt}

Next, we describe the discrete part of our optimization (green in~\cref{fig:lamp}). While continuous optimization can often successfully recover token embeddings close to the original, they can be in the wrong order, depending on how much positional embeddings influence the output. For example, reconstructions corresponding to sentences ``weather is nice.'' and ``nice weather is.'' might result in a similar reconstruction loss, though the first reconstruction has a higher likelihood of being natural text. To address this issue, we perform several discrete sequence transformations, and choose the one with both a low reconstruction loss and a low perplexity under the auxiliary language model.

\paragraph{Generating candidates}

Given the current reconstruction \mbox{$\vx = \vx_1 \vx_2 ... \vx_n$}, we generate candidates for the new reconstruction $\vx'$ using one of the following transformations:

\begin{itemize}
\item \emph{Swap}: We select two positions $i$ and $j$ in the sequence uniformly at random, and swap the tokens $\vx_i$ and $\vx_j$ at these two positions to obtain a new candidate sequence \mbox{$\vx' = \vx_1 \vx_2 \dots \vx_{i-1} \vx_j \vx_{i+1} \dots \vx_{j-1} \vx_i \vx_{j+1} \dots \vx_n$}.

\item \emph{MoveToken}: Similarly, we select two positions $i$ and $j$ in the sequence uniformly at random, and move the token $\vx_i$ after the position $j$ in the sequence, thus obtaining \mbox{$\vx' = \vx_1 \vx_2 \dots \vx_{i-1} \vx_{i+1} \dots \vx_{j-1} \vx_j \vx_i \vx_{j+1} \dots \vx_n$}.

\item \emph{MoveSubseq}: We select three positions $i$, $j$ and $p$ (where $i < j$) uniformly at random, and move the subsequence of tokens between $i$ and $j$ after position $p$. The new sequence is thus \mbox{$\vx' = \vx_1 \vx_2 \dots \vx_{i-1} \vx_{j+1} \dots \vx_p \vx_i \dots \vx_j \vx_{p+1} \dots \vx_n$}.

\item \emph{MovePrefix}: We select a position $i$ uniformly at random, and move the prefix of the sequence ending at position $i$ to the end of the sequence. The modified sequence then is \mbox{$\vx' = \vx_{i+1} \dots \vx_n \vx_1 \vx_2 \dots \vx_i$}.
\end{itemize}

Next, we use a language model to check if generated candidates improve over the current sequence.

\paragraph{Using a language model to select candidates}

We accept the new reconstruction $\vx'$ if it improves the combination of the reconstruction loss and perplexity:
\begin{equation*}
  \mathcal{L}_{\text{rec}}(\vx') + \alpha_{\text{lm}} \mathcal{L}_{\text{lm}}(\vt') < \mathcal{L}_{\text{rec}}(\vx) + \alpha_{\text{lm}} \mathcal{L}_{\text{lm}}(\vt)
\end{equation*}

\begin{wrapfigure}[21]{r}{0.48\textwidth}
    \vskip -0.340in
    \hfill
    \begin{minipage}{0.46\textwidth}
    \begin{algorithm}[H]
        \caption{Extracting text with LAMP}
        \label{alg:attack}
        \begin{algorithmic}[1]
          \STATE $\vx^{(k)} \sim \mathcal{N}(0, \mI), \text{where } k = 1,..., n_{\text{init}}$  \label{alg:line:initl}
          \STATE $\vx \leftarrow \argmin_{\vx^{(k)}} \mathcal{L}_{\text{rec}}(\vx^{(k)}) $
          \FOR{$i=1$ {\bfseries to} $it$} 
              \FOR{$j=1$ {\bfseries to} $n_c$} \label{alg:line:contopt}
              \STATE $\vx \gets \vx - \lambda \nabla_{\vx}  \mathcal{L}_{\text{rec}}(\vx)$ 
              \ENDFOR\label{alg:line:contoptend}
              \STATE $\vx_\text{best} \leftarrow \vx$ 
              \STATE $\vt_\text{best} \leftarrow \textsc{ProjectToVocab}(\vx_\text{best})$
              \STATE $L_\text{best} \leftarrow \mathcal{L}_{\text{rec}}(\vx_\text{best}) + \alpha_{\text{lm}} \mathcal{L}_{\text{lm}}(\vt_\text{best})$
              \FOR{$j=1$ {\bfseries to} $n_d$} \label{alg:line:discopt}
              \STATE $\vx' \leftarrow \textsc{Transform}(\vx)$ 
              \STATE $\vt' \leftarrow \textsc{ProjectToVocab}(\vx')$
              \STATE $L' \leftarrow  \mathcal{L}_{\text{rec}}(\vx') + \alpha_{\text{lm}} \mathcal{L}_{\text{lm}}(\vt')$
              \IF{$L' < L_\text{best}$}
                      \STATE $\vx_\text{best}, \vt_\text{best}, L_\text{best} \leftarrow \vx', \vt',L' $
                  \ENDIF
              \ENDFOR\label{alg:line:discoptend}
              \STATE $\vx \gets \vx_\text{best}$
          \ENDFOR
          \RETURN $\textsc{ProjectToVocab}(\vx)$ \label{alg:line:project}
        \end{algorithmic}
    \end{algorithm}

    \end{minipage}
  \end{wrapfigure}

Here $\vt$ and $\vt'$ denote sequences of tokens obtained by mapping each embedding of $\vx$ and $\vx'$ to the nearest neighbor in the vocabulary according to the cosine distance.
The term $\mathcal{L}_{\text{rec}}$ is the reconstruction loss introduced in~\cref{sec:contopt}, while $\mathcal{L}_{\text{lm}}$ denotes the perplexity measured by an auxiliary language model, such as GPT-2. The parameter $\alpha_{\text{lm}}$ determines the trade-off between $\mathcal{L}_{\text{rec}}$ and $\mathcal{L}_{\text{lm}}$: if it is too low then the attack will not utilize the language model, and if it is too high then the attack will disregard the reconstruction loss and only focus on the perplexity. Going back to our example, assume that our reconstruction equals the second sequence ``nice weather is.''. Then, at some point, we might use the \mbox{\emph{MoveToken}} transformation to move the word ``nice`` behind the word ``is'' which would presumably keep the reconstruction loss similar, but drastically improve perplexity.

\subsection{Complete Reconstruction Attack}

We present our end-to-end attack in~\cref{alg:attack}. We initialize the reconstruction $\vx$ by sampling from a Gaussian distribution $n_{\text{init}}$ times, and choose the sample with minimal reconstruction loss as our initial reconstruction. Then, at each step we alternate between continuous and discrete optimization. We first perform $n_c$ steps of continuous optimization to minimize the reconstruction loss (Lines~\ref{alg:line:contopt}-\ref{alg:line:contoptend}, see~\cref{sec:contopt}). Then, we perform $n_d$ steps of discrete optimization to minimize the combination of reconstruction loss and perplexity (Lines~\ref{alg:line:discopt}-\ref{alg:line:discoptend}, see~\cref{sec:discopt}). Finally, in Line~\ref{alg:line:project} we project the continuous embeddings $\vx$ to respective nearest tokens, according to cosine similarity.

\section{Experimental Evaluation} \label{sec:eval}

We now discuss our experimental results, demonstrating the effectiveness of \tool{} compared to prior work in a wide range of settings.
We present reconstruction results on several datasets, architectures, and batch sizes, together with the additional ablation study and evaluation of different defenses and training methods.

\paragraph{Datasets}
Prior work~\citep{tag} has demonstrated that text length is a key factor for the success of reconstruction from gradients. To this end, in our experiments we consider three binary classification datasets of increasing complexity: \cola{}~\cite{cola} and \sst{}~\cite{sst2} from GLUE~\cite{glue} with typical sequence lengths between 5 and 9 words, and 3 and 13 words, respectively, and \rotten{}~\cite{rotten} with typical sequence lengths between 14 and 27 words. The \cola{} dataset contains English sentences from language books annotated with binary labels describing if the sentences are grammatically correct, while \sst{} and \rotten{} contain movie reviews annotated with a binary sentiment. For all experiments, we evaluate the methods on 100 random sequences from the respective training sets.
We remark that attacking in binary classification setting is a more difficult task than in the masking setting considered by prior work~\citep{dlg}, where the attacker can utilize strictly more information.

\paragraph{Models} Our experiments are performed on different target models based on the BERT~\cite{bert} architecture. The main model we consider is \bertbase{}, which has $12$ layers, $768$ hidden units, $3072$ feed-forward filter size, and $12$ attention heads. To illustrate the generality of our approach with respect to model size, we additionally consider a larger model \bertlarge{}, which has $24$ layers, $1024$ hidden units, $4096$ feed-forward filter size, and $16$ attention heads as well as a smaller model \tinysix{} from \citet{tinybert} with $6$ layers, $768$ hidden units, feed-forward filter size of $3072$ and $12$ attention heads.
All models were taken from Hugging Face~\cite{hugging}. The \bertbase{} and \bertlarge{} were pretrained on Wikipedia~\cite{wikidump} and BookCorpus~\cite{Zhu_2015_ICCV} datasets, while \tinysix{} was distilled from \bertbase{}. We perform our main experiments on pretrained models, as this is the most common setting for training classification models from text~\cite{minaee2021deep}. For the auxiliary language model we use the pretrained GPT-2 provided by \citet{gdba}, trained on the same tokenizer used to pretrain our target BERT models.

\paragraph{Metrics} Following TAG~\cite{tag}, we measure the success of our methods based on the ROUGE family of metrics \cite{rouge}. In particular, we report the aggregated F-scores on ROUGE-1, ROUGE-2 and ROUGE-L, which measure the recovered unigrams, recovered bigrams and the ratio of the length of the longest matching subsequence to the length of whole sequence. When evaluating batch sizes greater than $1$, we exclude the padding tokens, used to pad shorter sequences, from the reconstruction and the ROUGE metric computation.

\paragraph{Experimental setup}
In all settings we consider, we compare our method with baselines DLG~\citep{dlg} and TAG~\citep{tag} discussed in~\cref{sec:related}.
As TAG does not have public code, we use our own implementation, and remark that the results obtained using our implementation are similar or better than those reported in \citet{tag}. We consider two variants of our attack, \lampcos{} and \lampell{}, that use the $\mathcal{L}_{\text{cos}}$ and $\mathcal{L}_{\text{tag}}$ gradient matching losses for the continuous optimization. For the \bertbase{} and \tinysix{} experiments, we run our attack with $it=30$, $n_c=75$ and $n_d=200$, and stop the optimization early once we reach a total of 2000 continuous optimization steps. For the \bertlarge{} model, whose number of parameters make the optimization harder, we use $it=25$ and $n_c=200$ instead, resulting in $5000$ continuous optimization steps. We run DLG and TAG for $10\,000$ optimization steps on \bertlarge{} and $2500$ on all other models. For the continuous optimization, we use Adam~\cite{adam} with a learning rate decay factor $\gamma$ applied every $50$ steps for all methods and experiments, except for \bertlarge{} ones where, following \citet{breaching}, we use AdamW~\cite{loshchilov2018decoupled} and linear learning rate decay schedule applied every step. We picked the hyperparameters for TAG, \lampcos{} and \lampell{}, separately on \cola{} and \rotten{} using grid search on \bertbase{} and applied them to all networks. As the optimal hyperparameters for \rotten{} exactly matched the ones on \cola{}, we used the same hyperparameters on \sst{}, as well. To account for the different optimizer used for \bertlarge{} models, we further tuned the learning rate $\lambda$ for \bertlarge{} experiments separately, keeping the other hyperparameters fixed. Additionally, for our methods we applied a two-step initialization procedure. We first initialized the embedding vectors with $500$ random samples from a standard Gaussian distribution and picked the best one according to $\mathcal{L}_{\text{grad}}(\vx)$. We then computed $500$ permutations on the best initialization and chose the best one in the same way. The effect of this procedure is investigated in \cref{app:inits}. Further details on our experimental setup are shown in \cref{app:addexpdet}.

\paragraph{Main experiments}
\begin{table*}[t]\centering
	
	\caption{Main results of text reconstruction from gradients with \tool{}, for various datasets and architectures in the setting with batch size equal to 1. FT denotes a fine-tuned model. R-1, R-2, and R-L, denote ROUGE-1, ROUGE-2 and ROUGE-L scores respectively.} \label{table:main}
	
	\newcommand{\threecol}[1]{\multicolumn{3}{c}{#1}}
	\newcommand{\fivecol}[1]{\multicolumn{5}{c}{#1}}
	\newcommand{\ninecol}[1]{\multicolumn{9}{c}{#1}}
	
	\newcommand{\bsz}{Batch Size~}
	\newcommand{\certified}{{CR(\%)}}
	
	\renewcommand{\arraystretch}{1.2}
	
	\newcommand{\ccellt}[2]{\colorbox{#1}{\makebox(20,8){{#2}}}}
	\newcommand{\ccellc}[2]{\colorbox{#1}{\makebox(8,8){{#2}}}}
	\newcommand{\ccells}[2]{\colorbox{#1}{\makebox(55,8){{#2}}}}
	
	\newcommand{\temp}[1]{\textcolor{red}{#1}}
	\newcommand{\noopcite}[1]{} 
	
	\newcommand{\skiplen}{0.004\linewidth} 
	\newcommand{\rlen}{0.01\linewidth} 
	
	\resizebox{0.95 \linewidth}{!}{
		\begingroup
		\setlength{\tabcolsep}{5pt} %
		\begin{tabular}{@{}c l rrr p{\skiplen}  rrr p{\skiplen} rrr p{\skiplen}  rrr p{\skiplen}  rrr  p{\skiplen}  rrr@{}} \toprule

			&& \threecol{\bertbase} && \threecol{\bertbase-FT} && \threecol{\tinysix} && \threecol{\bertlarge}\\
			
			\cmidrule(l{5pt}r{5pt}){3-5} \cmidrule(l{5pt}r{5pt}){7-9} \cmidrule(l{5pt}r{5pt}){11-13} \cmidrule(l{5pt}r{1pt}){15-17}
			
			&& R-1 & R-2 & R-L && R-1 & R-2 & R-L && R-1 & R-2 & R-L && R-1 & R-2 & R-L \\ \midrule
			\multirow{4}{*}{\cola}
			
			& DLG & 59.3 & 7.7 & 46.2 && 36.2 & 2.0 & 30.4 && 37.7 & 3.0 & 33.7 && 82.7 & 10.5 & 55.8\\
			& TAG & 78.9 & 10.2 & 53.3 && 40.2 & 3.1 & 32.3 &&  43.9 & 3.8 & 37.4 && 82.9 & 14.6 & 55.5 \\
			& \lampcos & {\bf 89.6}& {\bf 51.9}& {\bf 76.2} && {\bf 85.8}& {\bf 46.2}& {\bf 73.1} && 93.9 & {\bf 59.3}& {\bf 80.2} && {\bf 92.0} & {\bf 56.0} & {\bf 78.8}\\
			& \lampell & 87.5 & 47.5 & 73.2 && 40.3 & 9.3 & 35.2 && {\bf 94.5}& 52.1 & 76.0 && 91.2 & 47.8 & 75.4\\
			\midrule
			\multirow{4}{*}{\sst}
			
			& DLG & 65.4 & 17.7 & 54.2 && 36.0 & 2.7 & 33.9 && 42.0 & 5.4 & 39.6 && 78.4 & 18.1 & 59.0\\
			& TAG & 75.6 & 18.9 & 57.4 && 40.0 & 5.7 & 36.6 && 43.5 & 9.4 & 40.9 && 80.8 & 16.8 & 59.1\\ 
			& \lampcos & {\bf 88.8}& 56.9 & {\bf 77.7} && {\bf 87.6}& {\bf 54.1}& {\bf 76.1} && {\bf 91.6}& {\bf 58.2}& {\bf 79.7} && 88.5 & {\bf 55.9} & {\bf 76.5} \\
			& \lampell & 88.6 & {\bf 57.4}& 75.7 && 41.6 & 10.9 & 39.3 && 89.7 & 53.2 & 75.4 && {\bf 89.3} & 55.5 & 75.9\\
			\midrule      
			\multirow{4}{*}{\vtop{\hbox{\strut ~~Rotten}\hbox{\strut Tomatoes}}} 
			
			& DLG & 38.6 & 1.4 & 26.0 && 20.1 & 0.4 & 15.2 & &20.4 & 1.1 & 17.7 && 66.8 & 3.1 & 35.4\\
			& TAG & 60.3 & 3.5 & 33.6 && 26.7 & 0.9 & 18.2 && 25.8 & 1.5 & 20.2 && 73.6 & 4.4 & 36.8\\
			& \lampcos & {\bf 64.7}& {\bf 16.3}& {\bf 43.1} && {\bf 63.4}& {\bf 13.8}& {\bf 42.6} && {\bf 76.0}& {\bf 28.6}& {\bf 55.8} && 73.4 & 15.7 & 45.4\\
			&\lampell & 51.4 & 10.2 & 34.3 && 17.2 & 1.0 & 14.7 & & 74.0 & 19.4 & 46.7 && {\bf 77.6} & {\bf 16.6} & {\bf 45.8}\\
			\bottomrule
		\end{tabular}
		\endgroup
	}
\end{table*}

We evaluate the two variants of \tool{} against DLG~\citep{dlg} and TAG~\citep{tag} on \bertbase{}, \bertlarge{}, and \tinysix{}. Additionally, we evaluate attacks after \bertbase{} has already been fine-tuned for 2 epochs on each task (following~\citet{bert}), as \citet{bayesian} showed that in the vision domain it is significantly more difficult to attack already trained networks. For both baselines and our attacks, for simplicity we assume the lengths of sequences are known, as otherwise an adversary can simply run the attack for all possible lengths.
In the first experiment we consider setting where batch size is equal to 1.
The results are shown in~\cref{table:main}. From the ROUGE-1 metric, we can observe that we recover more tokens than the baselines in all settings.
Moreover, the main advantage of \tool{} is that the order of tokens in the reconstructed sequences matches the order in target sequences much more closely, as evidenced by the large increase in ROUGE-2 (5$\times$ on \cola{}). This observation is further backed by the ROUGE-L metric that shows we are on average able to reconstruct up to $23\%$ longer subsequences on the \bertbase{} model compared to the baselines.
These results confirm our intuition that guiding the search with GPT-2 allows us to reconstruct sequences that are a much closer match to the original sequences. We point out that~\cref{table:main} reaffirms the observations first made in \citet{tag}, that DLG is consistently worse in all metrics compared to both TAG and \tool{}, and that the significantly longer sequences in \rotten{} still pose challenges to good reconstruction.

Our results show that smaller and fine-tuned models also leak significant amount of client information. In particular, \tinysix{} is even more vulnerable than \bertbase{} and \bertbase{}-FT is shown to be only slightly worse in reconstruction compared to \bertbase{}, which is surprising given the prior image domain results. This shows that smaller models can not resolve the privacy issue, despite previous suggestions in \citet{tag}. Additionally, our \bertlarge{} experiments reaffirm the observation in \citet{tag} that the model is highly vulnerable to all attacks.

Further, we examine the variability of our \lampcos{} method with respect to random initialization. To this end, we ran the \bertbase{} experiment on \cola{} with 10 random seeds, which produced R-1, R-2 and R-L of $88.2 \pm 1.02$, $50.0 \pm 2.37$, $75.0 \pm 1.21$, respectively, which suggests that our results are consistent. Further, we assess the variability with respect to sentence choice in \cref{app:senteces}.

\begin{table*}[t]\centering
	
	\caption{Text reconstruction from gradients for different batch sizes $B$ on the \bertbase{} model. R-1, R-2, and R-L, denote ROUGE-1, ROUGE-2 and ROUGE-L scores respectively.} \label{table:batches}
	
	\newcommand{\threecol}[1]{\multicolumn{3}{c}{#1}} 
	\newcommand{\fivecol}[1]{\multicolumn{5}{c}{#1}}
	\newcommand{\ninecol}[1]{\multicolumn{9}{c}{#1}}
	
	\newcommand{\bsz}{Batch Size~}
	\newcommand{\certified}{{CR(\%)}}
	
	\renewcommand{\arraystretch}{1.2}
	
	\newcommand{\ccellt}[2]{\colorbox{#1}{\makebox(20,8){{#2}}}}
	\newcommand{\ccellc}[2]{\colorbox{#1}{\makebox(8,8){{#2}}}}
	\newcommand{\ccells}[2]{\colorbox{#1}{\makebox(55,8){{#2}}}}
	
	\newcommand{\temp}[1]{\textcolor{red}{#1}}
	\newcommand{\noopcite}[1]{} 
	
	\newcommand{\skiplen}{0.008\linewidth} 
	\newcommand{\rlen}{0.01\linewidth} 
	
	\resizebox{0.95 \linewidth}{!}{
		\begingroup 
		\setlength{\tabcolsep}{8pt} %
		\begin{tabular}{@{}c l rrr p{\skiplen}  rrr p{\skiplen} rrr p{\skiplen}  rrr p{\skiplen}  rrr  p{\skiplen}  rrr@{}} \toprule

			&& \threecol{$B$=1} && \threecol{$B$=2} && \threecol{$B$=4}\\
			
			\cmidrule(l{5pt}r{5pt}){3-5} \cmidrule(l{5pt}r{5pt}){7-9} \cmidrule(l{5pt}r{1pt}){11-13}
			
			&& R-1 & R-2 & R-L && R-1 & R-2 & R-L && R-1 & R-2 & R-L \\ \midrule
			\multirow{4}{*}{\cola}
			
			& DLG & 59.3 & 7.7 & 46.2 && 49.7 & 5.7 & 41.0 && 37.6 & 1.7 & 34.0 \\
			& TAG & 78.9 & 10.2 & 53.3 && 68.8 & 7.6 & 49.0 && 56.2 & 6.7 & 44.0 \\
			& \lampcos & {\bf 89.6}& {\bf 51.9}& {\bf 76.2} && 74.4 & 29.5 & 61.9 && 55.2 & 14.5 & 48.0 \\
			& \lampell & 87.5 & 47.5 & 73.2 && {\bf 78.0}& {\bf 31.4}& {\bf 63.7}&& {\bf 66.2}& {\bf 21.8}& {\bf 55.2}\\
			
			\midrule
			\multirow{4}{*}{\sst}
			
			& DLG & 65.4 & 17.7 & 54.2 && 57.7 & 11.7 & 48.2 && 43.1 & 6.8 & 39.4\\
			& TAG & 75.6 & 18.9 & 57.4 && 71.8 & 16.1 & 54.4 && 61.0 & 12.3 & 48.4\\ 
			& \lampcos & {\bf 88.8}& 56.9 & {\bf 77.7} && 72.2 & 37.0 & 63.6 && 57.9 & 23.4 & 52.3 \\
			& \lampell & 88.6 & {\bf 57.4}& 75.7 && {\bf 82.5}& {\bf 45.8}& {\bf 70.8}&& {\bf{69.5}}& {\bf{32.5}}& {\bf{59.9}}\\
			
			\midrule      
			\multirow{4}{*}{\vtop{\hbox{\strut ~~Rotten}\hbox{\strut Tomatoes}}} 
			
			& DLG & 38.6 & 1.4 & 26.0 && 29.2 & 1.1 & 23.0 && 21.2 & 0.5 & 18.6\\
			& TAG & 60.3 & 3.5 & 33.6 && {\bf 47.4}& 2.7 & 29.5 && 32.3 & 1.4 & 23.5\\
			& \lampcos & {\bf 64.7}& {\bf 16.3}& {\bf 43.1} && 37.4 & 5.6 & 29.0 && 25.7 & 1.8 & 22.1\\
			&\lampell & 51.4 & 10.2 & 34.3 && 46.3 & {\bf 7.6} & {\bf 32.7} && {\bf 35.1}& {\bf 4.2} & {\bf 27.2}\\

			\bottomrule
		\end{tabular}
		\endgroup
	}
\end{table*}

\paragraph{Larger batch sizes} Unlike prior work, we also evaluate the different attacks on updates computed on batch sizes greater than 1 on the \bertbase{} model to investigate whether we can reconstruct some sequences in this more challenging setting. The results are shown in \cref{table:batches}. Similarly to the results in \cref{table:main}, we observe that we obtain better results than the baselines on all ROUGE metrics in all experiments, except on \rotten{} with batch size 2, where TAG obtains slightly better ROGUE-1. Our experiments show that for larger batch sizes we can also reconstruct significant portions of text (see experiments on \cola{} and \sst{}). To the best of our knowledge, we are the first to show this, suggesting that gradient leakage can be a realistic security threat in practice. Comparing the results for \lampell{} and \lampcos{}, we observe that $\mathcal{L}_{\text{cos}}$ is better than $\mathcal{L}_{\text{tag}}$ in almost all metrics on batch size $1$, across models, but the trend reverses as batch size is increased.

\definecolor{refcolor}{rgb}{0.15, 0.38, 0.61}
\definecolor{bigramcolor}{rgb}{0.03, 0.47, 0.19}
\definecolor{unigramcolor}{rgb}{0.85, 0.57, 0.0}

\definecolor{bigramcolor}{rgb}{0.7, 0.93, 0.36}
\definecolor{unigramcolor}{rgb}{0.99, 0.97, 0.37}
    
 \newcommand{\unigram}[1]{\colorbox{unigramcolor}{\vrule height4.2pt depth1pt width0pt #1}}
 \newcommand{\bigram}[1]{\colorbox{bigramcolor}{\vrule height4.2pt depth1pt width0pt #1}}

\begin{table*}[t]\centering  
    \caption{The result of text reconstruction on several examples from the dataset (for \bertbase~with $B$=1). We show only TAG (better baseline) and \lampcos~as it is superior in these cases.} \label{table:sentences}
  
    \newcommand{\threecol}[1]{\multicolumn{3}{c}{#1}}
    \newcommand{\fivecol}[1]{\multicolumn{5}{c}{#1}}
    \newcommand{\ninecol}[1]{\multicolumn{9}{c}{#1}}
  
    \newcommand{\bsz}{Batch Size~}
   
    \renewcommand{\arraystretch}{1.4}
  
    \newcommand{\ccellt}[2]{\colorbox{#1}{\makebox(20,8){{#2}}}}
    \newcommand{\ccellc}[2]{\colorbox{#1}{\makebox(8,8){{#2}}}}
    \newcommand{\ccells}[2]{\colorbox{#1}{\makebox(55,8){{#2}}}}
  
    \newcommand{\noopcite}[1]{} 
    
    \resizebox{0.95\linewidth}{!}{
      \begin{tabular}{@{}c l p{13cm}@{}} \toprule
  
      & & Sequence \\ \midrule
    \multirow{3}{*}{\cola} 
    & \textcolor{refcolor}{Reference} & \textcolor{refcolor}{mary has never kissed a man who is taller than john.} \\ 
        & TAG & \unigram{man} seem \bigram{taller than} \unigram{mary},. \unigram{kissed} has \unigram{john} mph \unigram{never} \\ 
        & \lampcos &  \bigram{mary has never kissed a man who is taller than john.} \\ \cmidrule{2-3}

    \multirow{3}{*}{\sst}
    & \textcolor{refcolor}{Reference} & \textcolor{refcolor}{i also believe that resident evil is not it.} \\ 
        & TAG & \unigram{resident}. or. \unigram{is} pack down \unigram{believe} \unigram{i} \unigram{evil}  \\ 
        & \lampcos &  \bigram{i also believe that resident} \bigram{resident evil} \bigram{not it}. \\ \cmidrule{2-3} 
        
    \multirow{3}{*}{\vtop{\hbox{\strut ~~Rotten}\hbox{\strut Tomatoes}}} 
    & \textcolor{refcolor}{Reference}  & \textcolor{refcolor}{a well - made and often lovely depiction of the mysteries of friendship.} \\ 
    & TAG & - \unigram{the} \unigram{friendship} taken \unigram{and} \unigram{lovely} \unigram{a} \unigram{made} \unigram{often} \bigram{depiction of} \unigram{well} \unigram{mysteries}. \\ 
    & \lampcos & \bigram{a well} \unigram{often} \unigram{made} - \unigram{and} \bigram{lovely depiction} \bigram{mysteries of} \bigram{mysteries of friendship}. \\ 
    
      \bottomrule
      \end{tabular}}
  \end{table*}

\paragraph{Sample reconstructions} We show sample sequence reconstructions from both \tool{} and the TAG baseline on CoLA with $B=1$ in~\cref{table:sentences}, marking the correctly reconstructed bigrams with green and correct unigrams with yellow. We can observe that our reconstruction is more coherent, and that it qualitatively outperforms the baseline. In \cref{app:stepbystep}, we show the convergence rate of our method compared to the baselines on an example sequence, suggesting that LAMP can often converges faster.

\paragraph{Ablation studies}
\begin{table}[t]\centering
  \tiny 

  \caption{An ablation study with the \bertbase~($B$=1) model. We restate the results for \lampcos~and \lampell~from \cref{table:main} and introduce four ablations, done on the better of the two variants of \tool, in these cases \lampcos.} \label{table:ablation}
 
  \newcommand{\threecol}[1]{\multicolumn{3}{c}{#1}}
  \newcommand{\fivecol}[1]{\multicolumn{5}{c}{#1}}
  \newcommand{\ninecol}[1]{\multicolumn{9}{c}{#1}}

  \newcommand{\bsz}{Batch Size~}

  \renewcommand{\arraystretch}{1.2}

  \newcommand{\ccellt}[2]{\colorbox{#1}{\makebox(20,8){{#2}}}}
  \newcommand{\ccellc}[2]{\colorbox{#1}{\makebox(8,8){{#2}}}}
  \newcommand{\ccells}[2]{\colorbox{#1}{\makebox(55,8){{#2}}}}

  \newcommand{\noopcite}[1]{}
  \newcommand{\skiplen}{0.001\linewidth}  
  
  \resizebox{0.95 \columnwidth}{!}{
    \begin{tabular}{@{}l rrr p{\skiplen} rrr p{\skiplen} rrr@{}} \toprule

      & \threecol{\cola} && \threecol{\sst} && \threecol{Rotten Tomatoes} \\
      \cmidrule(l{5pt}r{5pt}){2-4} \cmidrule(l{5pt}r{5pt}){6-8} \cmidrule(l{5pt}r{5pt}){10-12}
      & R-1 & R-2 & R-L && R-1 & R-2 & R-L && R-1 & R-2 & R-L  \\      
      \midrule
   
    \lampcos & {\bf 89.6}& {\bf 51.9}& {\bf 76.2} && {\bf 88.8}& 56.9 & {\bf 77.7} && 64.7& {\bf 16.3}& {\bf 43.1} \\
    \lampell & 87.5 & 47.5 & 73.2 && 88.6 & {\bf 57.4}& 75.7 && 51.4 & 10.2 & 34.3  \\
     \cmidrule(l{0pt}r{0pt}){1-12} 
     \lampltwo & 69.4 & 30.1 & 58.8 && 72.4 & 44.1 & 65.4 && 31.9 & 5.5 & 25.7  \\
     \lampnoperp &  86.7 & 26.6 & 66.9 &&  82.6 & 37.0 & 68.4 &&  64.0 & 9.9 & 40.3  \\
     \lampnoreg & 84.5 & 38.0 & 69.1 && 83.3 & 44.7 & 71.9 && 57.8 & 11.1 & 38.3  \\
     \lampdiscafter & 87.4 & 28.6 & 66.9 && 85.4 & 42.4 & 71.0 && {\bf 65.0} & 11.4 & 42.3 \\
     \lampnoswaps & 86.6 & 29.6 & 67.4 && 84.1 & 40.0 & 70.0 && 61.5 & 10.2 & 40.8  \\

     \bottomrule
     \end{tabular}
  }
\end{table}

In the next experiment, we perform ablation studies to  examine the influence of each proposed component of our method.
We compare the following variants of LAMP: (i) with cosine loss, (ii) with $L_1$ + $L_2$ loss, (iii) with $L_2$ loss,  (iv) without the language model ($\alpha_{\text{lm}}=0$), (v) without embedding regularization ($\alpha_{\text{reg}}=0$), (vi) without alternating of the discrete and continuous optimization steps---executing $it \cdot n_c$ continuous optimization steps first, followed by $it$ discrete optimizations with $n_d$ steps each, (vii) without discrete transformations ($n_d = 0$).
For this experiment, we use the \cola{} dataset and \bertbase{} with $B=1$.
We show the results in~\cref{table:ablation}.
We observe that \tool{} achieves good results with both losses, though cosine is generally better for batch size $1$.
More importantly, dropping any of the proposed features makes ROUGE-1 and ROUGE-2 significantly worse. We note the most significant drop in ROUGE-2 reconstruction quality happens when using transformations without using the language model~(\lampnoperp), which performs even worse than doing no transformations~(\lampnoswaps) at all. This suggests that the use of the language model is crucial to obtaining good results.
Further, we observe that our proposed scheme for alternating the continuous and discrete optimization steps is important, as doing the discrete optimization at the end~(\lampdiscafter) for the same number of steps results in reconstructions only marginally better (in ROUGE-2) compared to the reconstructions obtained without any discrete optimization~(\lampnoswaps). The experiments also confirm usefulness of other features such as embedding regularization.

\paragraph{Attacking defended networks}

\begin{table}[t]\centering
  \caption{Evaluation on gradients defended with Gaussian noise, with \bertbase~($B$=1) on the \cola~dataset.} \label{table:defended}

  \newcommand{\threecol}[1]{\multicolumn{3}{c}{#1}}
  \newcommand{\fivecol}[1]{\multicolumn{5}{c}{#1}}
  \newcommand{\ninecol}[1]{\multicolumn{9}{c}{#1}}

  \newcommand{\bsz}{Batch Size~}

  \renewcommand{\arraystretch}{1.2}

  \newcommand{\ccellt}[2]{\colorbox{#1}{\makebox(20,8){{#2}}}}
  \newcommand{\ccellc}[2]{\colorbox{#1}{\makebox(8,8){{#2}}}}
  \newcommand{\ccells}[2]{\colorbox{#1}{\makebox(55,8){{#2}}}}

  \newcommand{\noopcite}[1]{} 
  \newcommand{\skiplen}{0.01\linewidth} 
  
  \resizebox{ 0.95 \columnwidth}{!}{
    \begin{tabular}{@{}l rrr p{\skiplen} rrr p{\skiplen} rrr p{\skiplen} rrr@{}} \toprule

      & \threecol{$\sigma = 0.001$} && \threecol{$\sigma = 0.002$} && \threecol{$\sigma = 0.005$} && \threecol{$\sigma = 0.01$} \\
      & \threecol{MCC$ =0.551$} && \threecol{MCC$ =0.526$} && \threecol{MCC$ =0.464$} && \threecol{MCC$ =0.364$} \\
 
   \cmidrule(l{5pt}r{5pt}){2-4} \cmidrule(l{5pt}r{2pt}){6-8} \cmidrule(l{5pt}r{2pt}){10-12} \cmidrule(l{5pt}r{2pt}){14-16}
 
    & R-1 & R-2 & R-L && R-1 & R-2 & R-L && R-1 & R-2 & R-L && R-1 & R-2 & R-L \\ \midrule
     DLG & 60.0 & 7.2 & 46.3 && 61.3 & 7.5 & 47.0 && 58.8 & 8.0 & 46.4 && 56.4 & 6.3 & 44.8 \\
    TAG & 70.7 & 6.0 & 50.8 && 67.1 & 8.4 & 49.9 && 64.1 & 6.5 & 47.6 && 59.6 & 6.5 & 46.2 \\
    \lampcos & {\bf 81.2}& {\bf 42.7}& {\bf 69.4}&& 70.6 & 29.5 & 60.9 && 43.3 & 9.45 & 39.7 && 27.7 & 2.0 & 27.6 \\
    \lampell & 79.2 & 32.8 & 64.1 && {\bf 74.3}& {\bf 31.0}& {\bf 61.9} && \textbf{73.5} & \textbf{29.7} & \textbf{60.9} && \textbf{69.6} & \textbf{29.4} & \textbf{60.6}\\
     \bottomrule
     \end{tabular}
  }
\end{table}

So far, all experiments assumed that clients have not defended against data leakage. Following work on vision attacks~\cite{dlg,wei2020framework}, we now consider the defense of adding Gaussian noise to gradients (with additional clipping this would correspond to DP-SGD~\citep{dpsgd}). Note that, as usual, there is a trade-off between privacy and accuracy: adding more noise will lead to better privacy, but make accuracy worse. We measure the performance of the fine-tuned models on \cola{} using the MCC metric~\citep{mcc} for which higher values are better. The fine-tuning was done for 2 epochs with different Gaussian noise levels $\sigma$, and we obtained the MCC scores depicted in ~\cref{table:defended}. We did not explore noises $>0.01$ due to the significant drop in MCC from $0.557$ for the undefended model to $0.364$. The results of our experiments on these defended networks are presented in~\cref{table:defended}. While all methods' reconstruction metrics degrade, as expected, we see that most text is still recoverable for the chosen noise levels. Moreover, our method still outperforms the baselines, and thus shows the importance of evaluating defenses with strong reconstruction attacks.
In~\cref{app:masking} we show that \tool{} is also useful against a defense which masks some percentage of gradients.

\paragraph{Attacking FedAvg}

So far, we have only considered attacking the FedSGD algorithm. In this experiment, we apply our attack on the commonly used FedAvg~\cite{konevcny2016federated} algorithm. As NLP models are often fine-tuned using small learning rates (2e-5 to 5e-5 in the original BERT paper), we find that FedAvg reconstruction performance is close to FedSGD performance with batch size multiplied by the number of FedAvg steps. We experimented with attacking FedAvg with 4 steps using $B=1$ per step, $lr=5\text{e-}5$ on CoLA and \bertbase{} with \lampell{}. We obtained R-1, R-2 and R-L of $66.5$, $21.0$, $55.1$, respectively, comparable to the reported results on FedSGD with $B=4$.

\section{Conclusion} \label{sec:conclusion}
In this paper, we presented \tool{}, a new method for reconstructing private text data from gradients by leveraging language model priors and alternating discrete and continuous optimization. Our extensive experimental evaluation showed that \tool{} consistently outperforms prior work on datasets of varying complexity and models of different sizes. Further, we established that \tool{} is able to reconstruct private data in a number of challenging settings, including bigger batch sizes, noise-defended gradients, and fine-tuned models. Our work highlights that private text data is not sufficiently protected by federated learning algorithms and that more work is needed to alleviate this issue.

\message{^^JLASTBODYPAGE \thepage^^J}

\bibliography{references}

\begin{thebibliography}{44}
\providecommand{\natexlab}[1]{#1}
\providecommand{\url}[1]{\texttt{#1}}
\expandafter\ifx\csname urlstyle\endcsname\relax
  \providecommand{\doi}[1]{doi: #1}\else
  \providecommand{\doi}{doi: \begingroup \urlstyle{rm}\Url}\fi

\bibitem[Abadi et~al.(2016)Abadi, Chu, Goodfellow, McMahan, Mironov, Talwar,
  and Zhang]{dpsgd}
Martin Abadi, Andy Chu, Ian Goodfellow, H.~Brendan McMahan, Ilya Mironov, Kunal
  Talwar, and Li~Zhang.
\newblock Deep learning with differential privacy.
\newblock \emph{Proceedings of the 2016 ACM SIGSAC Conference on Computer and
  Communications Security (ACM CCS), pp. 308-318, 2016}, 2016.
\newblock \doi{10.1145/2976749.2978318}.

\bibitem[Balunovi{\'c} et~al.(2021)Balunovi{\'c}, Dimitrov, Staab, and
  Vechev]{bayesian}
Mislav Balunovi{\'c}, Dimitar~I Dimitrov, Robin Staab, and Martin Vechev.
\newblock Bayesian framework for gradient leakage.
\newblock \emph{arXiv preprint arXiv:2111.04706}, 2021.

\bibitem[Deng et~al.(2021)Deng, Wang, Li, Wang, Shang, Liu, Rajasekaran, and
  Ding]{tag}
Jieren Deng, Yijue Wang, Ji~Li, Chenghong Wang, Chao Shang, Hang Liu,
  Sanguthevar Rajasekaran, and Caiwen Ding.
\newblock {TAG}: Gradient attack on transformer-based language models.
\newblock In \emph{Findings of the Association for Computational Linguistics:
  EMNLP 2021}, pages 3600--3610, Punta Cana, Dominican Republic, November 2021.
  Association for Computational Linguistics.
\newblock \doi{10.18653/v1/2021.findings-emnlp.305}.

\bibitem[Devlin et~al.(2018)Devlin, Chang, Lee, and Toutanova]{bert}
Jacob Devlin, Ming-Wei Chang, Kenton Lee, and Kristina Toutanova.
\newblock Bert: Pre-training of deep bidirectional transformers for language
  understanding.
\newblock \emph{arXiv preprint arXiv:1810.04805}, 2018.

\bibitem[Foundation()]{wikidump}
Wikimedia Foundation.
\newblock Wikimedia downloads.
\newblock URL \url{https://dumps.wikimedia.org}.

\bibitem[Fowl et~al.(2022)Fowl, Geiping, Reich, Wen, Czaja, Goldblum, and
  Goldstein]{fowl2022decepticons}
Liam Fowl, Jonas Geiping, Steven Reich, Yuxin Wen, Wojtek Czaja, Micah
  Goldblum, and Tom Goldstein.
\newblock Decepticons: Corrupted transformers breach privacy in federated
  learning for language models.
\newblock \emph{arXiv preprint arXiv:2201.12675}, 2022.

\bibitem[Geiping et~al.(2020)Geiping, Bauermeister, Dr\"{o}ge, and
  Moeller]{geiping}
Jonas Geiping, Hartmut Bauermeister, Hannah Dr\"{o}ge, and Michael Moeller.
\newblock Inverting gradients - how easy is it to break privacy in federated
  learning?
\newblock In H.~Larochelle, M.~Ranzato, R.~Hadsell, M.~F. Balcan, and H.~Lin,
  editors, \emph{Advances in Neural Information Processing Systems}, volume~33,
  pages 16937--16947. Curran Associates, Inc., 2020.

\bibitem[Geiping et~al.(2022)Geiping, Fowl, and Wen]{breaching}
Jonas Geiping, Liam Fowl, and Yuxin Wen.
\newblock Breaching - a framework for attacks against privacy in federated
  learning.
\newblock 2022.
\newblock URL \url{https://github.com/JonasGeiping/breaching}.

\bibitem[Geng et~al.(2021)Geng, Mou, Li, Li, Beyan, Decker, and
  Rong]{geng2021towards}
Jiahui Geng, Yongli Mou, Feifei Li, Qing Li, Oya Beyan, Stefan Decker, and
  Chunming Rong.
\newblock Towards general deep leakage in federated learning.
\newblock \emph{arXiv preprint arXiv:2110.09074}, 2021.

\bibitem[Guo et~al.(2021)Guo, Sablayrolles, J{\'e}gou, and Kiela]{gdba}
Chuan Guo, Alexandre Sablayrolles, Herv{\'e} J{\'e}gou, and Douwe Kiela.
\newblock Gradient-based adversarial attacks against text transformers.
\newblock \emph{arXiv preprint arXiv:2104.13733}, 2021.

\bibitem[Gupta et~al.(2022)Gupta, Huang, Zhong, Gao, Li, and
  Chen]{gupta2022recovering}
Samyak Gupta, Yangsibo Huang, Zexuan Zhong, Tianyu Gao, Kai Li, and Danqi Chen.
\newblock Recovering private text in federated learning of language models.
\newblock \emph{arXiv preprint arXiv:2205.08514}, 2022.

\bibitem[Huang et~al.(2021)Huang, Gupta, Song, Li, and Arora]{arora}
Yangsibo Huang, Samyak Gupta, Zhao Song, Kai Li, and Sanjeev Arora.
\newblock Evaluating gradient inversion attacks and defenses in federated
  learning.
\newblock \emph{Advances in Neural Information Processing Systems}, 34, 2021.

\bibitem[Jelinek et~al.(1977)Jelinek, Mercer, Bahl, and
  Baker]{jelinek1977perplexity}
Fred Jelinek, Robert~L Mercer, Lalit~R Bahl, and James~K Baker.
\newblock Perplexity—a measure of the difficulty of speech recognition tasks.
\newblock \emph{The Journal of the Acoustical Society of America}, 62\penalty0
  (S1):\penalty0 S63--S63, 1977.

\bibitem[Jeon et~al.(2021)Jeon, Lee, Oh, Ok, et~al.]{jeon2021gradient}
Jiwnoo Jeon, Kangwook Lee, Sewoong Oh, Jungseul Ok, et~al.
\newblock Gradient inversion with generative image prior.
\newblock \emph{Advances in Neural Information Processing Systems}, 34, 2021.

\bibitem[Jiao et~al.(2019)Jiao, Yin, Shang, Jiang, Chen, Li, Wang, and
  Liu]{tinybert}
Xiaoqi Jiao, Yichun Yin, Lifeng Shang, Xin Jiang, Xiao Chen, Linlin Li, Fang
  Wang, and Qun Liu.
\newblock Tinybert: Distilling bert for natural language understanding.
\newblock \emph{arXiv preprint arXiv:1909.10351}, 2019.

\bibitem[Kairouz et~al.(2019)Kairouz, McMahan, Avent, Bellet, Bennis, Bhagoji,
  Bonawitz, Charles, Cormode, Cummings, et~al.]{advances}
Peter Kairouz, H~Brendan McMahan, Brendan Avent, Aur{\'e}lien Bellet, Mehdi
  Bennis, Arjun~Nitin Bhagoji, Keith Bonawitz, Zachary Charles, Graham Cormode,
  Rachel Cummings, et~al.
\newblock Advances and open problems in federated learning.
\newblock \emph{arXiv preprint arXiv:1912.04977}, 2019.

\bibitem[Kingma and Ba(2014)]{adam}
Diederik~P Kingma and Jimmy Ba.
\newblock Adam: A method for stochastic optimization.
\newblock \emph{arXiv preprint arXiv:1412.6980}, 2014.

\bibitem[Kone{\v{c}}n{\'y} et~al.(2016)Kone{\v{c}}n{\'y}, McMahan, Ramage, and
  Richt{\'{a}}rik]{fedopt}
Jakub Kone{\v{c}}n{\'y}, H.~Brendan McMahan, Daniel Ramage, and Peter
  Richt{\'{a}}rik.
\newblock Federated optimization: Distributed machine learning for on-device
  intelligence.
\newblock \emph{arXiv preprint arXiv:1610.02527}, 2016.

\bibitem[Kone{\v{c}}n{\`y} et~al.(2016)Kone{\v{c}}n{\`y}, McMahan, Yu,
  Richt{\'a}rik, Suresh, and Bacon]{konevcny2016federated}
Jakub Kone{\v{c}}n{\`y}, H~Brendan McMahan, Felix~X Yu, Peter Richt{\'a}rik,
  Ananda~Theertha Suresh, and Dave Bacon.
\newblock Federated learning: Strategies for improving communication
  efficiency.
\newblock \emph{arXiv preprint arXiv:1610.05492}, 2016.

\bibitem[Lin(2004)]{rouge}
Chin-Yew Lin.
\newblock Rouge: A package for automatic evaluation of summaries.
\newblock In \emph{Text summarization branches out}, pages 74--81, 2004.

\bibitem[Loshchilov and Hutter(2019)]{loshchilov2018decoupled}
Ilya Loshchilov and Frank Hutter.
\newblock Decoupled weight decay regularization.
\newblock In \emph{International Conference on Learning Representations}, 2019.
\newblock URL \url{https://openreview.net/forum?id=Bkg6RiCqY7}.

\bibitem[Lu et~al.(2021)Lu, Zhang, Zhao, He, and Cheng]{april}
Jiahao Lu, Xi~Sheryl Zhang, Tianli Zhao, Xiangyu He, and Jian Cheng.
\newblock April: Finding the achilles' heel on privacy for vision transformers.
\newblock \emph{arXiv preprint arXiv:2112.14087}, 2021.

\bibitem[Matthews(1975)]{mcc}
Brian~W Matthews.
\newblock Comparison of the predicted and observed secondary structure of t4
  phage lysozyme.
\newblock \emph{Biochimica et Biophysica Acta (BBA)-Protein Structure},
  405\penalty0 (2):\penalty0 442--451, 1975.

\bibitem[McMahan et~al.(2017)McMahan, Moore, Ramage, Hampson, and
  y~Arcas]{fedsgd}
Brendan McMahan, Eider Moore, Daniel Ramage, Seth Hampson, and
  Blaise~Ag{\"{u}}era y~Arcas.
\newblock Communication-efficient learning of deep networks from decentralized
  data.
\newblock In \emph{AISTATS}, 2017.

\bibitem[Minaee et~al.(2021)Minaee, Kalchbrenner, Cambria, Nikzad, Chenaghlu,
  and Gao]{minaee2021deep}
Shervin Minaee, Nal Kalchbrenner, Erik Cambria, Narjes Nikzad, Meysam
  Chenaghlu, and Jianfeng Gao.
\newblock Deep learning--based text classification: A comprehensive review.
\newblock \emph{ACM Computing Surveys (CSUR)}, 54\penalty0 (3):\penalty0 1--40,
  2021.

\bibitem[Pang and Lee(2005)]{rotten}
Bo~Pang and Lillian Lee.
\newblock Seeing stars: Exploiting class relationships for sentiment
  categorization with respect to rating scales.
\newblock In \emph{Proceedings of the ACL}, 2005.

\bibitem[Paszke et~al.(2019)Paszke, Gross, Massa, Lerer, Bradbury, Chanan,
  Killeen, Lin, Gimelshein, Antiga, Desmaison, Kopf, Yang, DeVito, Raison,
  Tejani, Chilamkurthy, Steiner, Fang, Bai, and Chintala]{NEURIPS2019_9015}
Adam Paszke, Sam Gross, Francisco Massa, Adam Lerer, James Bradbury, Gregory
  Chanan, Trevor Killeen, Zeming Lin, Natalia Gimelshein, Luca Antiga, Alban
  Desmaison, Andreas Kopf, Edward Yang, Zachary DeVito, Martin Raison, Alykhan
  Tejani, Sasank Chilamkurthy, Benoit Steiner, Lu~Fang, Junjie Bai, and Soumith
  Chintala.
\newblock Pytorch: An imperative style, high-performance deep learning library.
\newblock In H.~Wallach, H.~Larochelle, A.~Beygelzimer, F.~d\textquotesingle
  Alch\'{e}-Buc, E.~Fox, and R.~Garnett, editors, \emph{Advances in Neural
  Information Processing Systems 32}, pages 8024--8035. Curran Associates,
  Inc., 2019.
\newblock URL
  \url{http://papers.neurips.cc/paper/9015-pytorch-an-imperative-style-high-performance-deep-learning-library.pdf}.

\bibitem[Phong et~al.(2017)Phong, Aono, Hayashi, Wang, and
  Moriai]{phong2017privacy}
Le~Trieu Phong, Yoshinori Aono, Takuya Hayashi, Lihua Wang, and Shiho Moriai.
\newblock Privacy-preserving deep learning: Revisited and enhanced.
\newblock In \emph{{ATIS}}, volume 719 of \emph{Communications in Computer and
  Information Science}, pages 100--110. Springer, 2017.

\bibitem[Radford et~al.(2019)Radford, Wu, Child, Luan, Amodei, Sutskever,
  et~al.]{radford2019gpt2}
Alec Radford, Jeffrey Wu, Rewon Child, David Luan, Dario Amodei, Ilya
  Sutskever, et~al.
\newblock Language models are unsupervised multitask learners.
\newblock \emph{OpenAI blog}, 1\penalty0 (8):\penalty0 9, 2019.

\bibitem[Ramaswamy et~al.(2019)Ramaswamy, Mathews, Rao, and
  Beaufays]{ramaswamy2019keyboard}
Swaroop Ramaswamy, Rajiv Mathews, Kanishka Rao, and Fran{\c{c}}oise Beaufays.
\newblock Federated learning for emoji prediction in a mobile keyboard.
\newblock \emph{CoRR}, abs/1906.04329, 2019.

\bibitem[Scheliga et~al.(2021)Scheliga, Mäder, and Seeland]{precode}
Daniel Scheliga, Patrick Mäder, and Marco Seeland.
\newblock Precode - a generic model extension to prevent deep gradient leakage,
  2021.

\bibitem[Shejwalkar et~al.(2021)Shejwalkar, Houmansadr, Kairouz, and
  Ramage]{shejwalkar2021back}
Virat Shejwalkar, Amir Houmansadr, Peter Kairouz, and Daniel Ramage.
\newblock Back to the drawing board: A critical evaluation of poisoning attacks
  on federated learning.
\newblock \emph{arXiv preprint arXiv:2108.10241}, 2021.

\bibitem[Socher et~al.(2013)Socher, Perelygin, Wu, Chuang, Manning, Ng, and
  Potts]{sst2}
Richard Socher, Alex Perelygin, Jean Wu, Jason Chuang, Christopher~D Manning,
  Andrew~Y Ng, and Christopher Potts.
\newblock Recursive deep models for semantic compositionality over a sentiment
  treebank.
\newblock In \emph{Proceedings of the 2013 conference on empirical methods in
  natural language processing}, pages 1631--1642, 2013.

\bibitem[Sun et~al.(2021)Sun, Li, Wang, Yang, Li, and Chen]{soteria}
Jingwei Sun, Ang Li, Binghui Wang, Huanrui Yang, Hai Li, and Yiran Chen.
\newblock Soteria: Provable defense against privacy leakage in federated
  learning from representation perspective.
\newblock In \emph{{CVPR}}, pages 9311--9319. Computer Vision Foundation /
  {IEEE}, 2021.

\bibitem[Vaswani et~al.(2017)Vaswani, Shazeer, Parmar, Uszkoreit, Jones, Gomez,
  Kaiser, and Polosukhin]{vaswani2017attention}
Ashish Vaswani, Noam Shazeer, Niki Parmar, Jakob Uszkoreit, Llion Jones,
  Aidan~N Gomez, {\L}ukasz Kaiser, and Illia Polosukhin.
\newblock Attention is all you need.
\newblock In \emph{Advances in neural information processing systems}, pages
  5998--6008, 2017.

\bibitem[Wang et~al.(2018)Wang, Singh, Michael, Hill, Levy, and Bowman]{glue}
Alex Wang, Amanpreet Singh, Julian Michael, Felix Hill, Omer Levy, and Samuel~R
  Bowman.
\newblock Glue: A multi-task benchmark and analysis platform for natural
  language understanding.
\newblock \emph{arXiv preprint arXiv:1804.07461}, 2018.

\bibitem[Warstadt et~al.(2019)Warstadt, Singh, and Bowman]{cola}
Alex Warstadt, Amanpreet Singh, and Samuel~R Bowman.
\newblock Neural network acceptability judgments.
\newblock \emph{Transactions of the Association for Computational Linguistics},
  7:\penalty0 625--641, 2019.

\bibitem[Wei et~al.(2020)Wei, Liu, Loper, Chow, Gursoy, Truex, and
  Wu]{wei2020framework}
Wenqi Wei, Ling Liu, Margaret Loper, Ka-Ho Chow, Mehmet~Emre Gursoy, Stacey
  Truex, and Yanzhao Wu.
\newblock A framework for evaluating gradient leakage attacks in federated
  learning.
\newblock \emph{arXiv preprint arXiv:2004.10397}, 2020.

\bibitem[Wolf et~al.(2020)Wolf, Debut, Sanh, Chaumond, Delangue, Moi, Cistac,
  Rault, Louf, Funtowicz, Davison, Shleifer, von Platen, Ma, Jernite, Plu, Xu,
  Scao, Gugger, Drame, Lhoest, and Rush]{hugging}
Thomas Wolf, Lysandre Debut, Victor Sanh, Julien Chaumond, Clement Delangue,
  Anthony Moi, Pierric Cistac, Tim Rault, Rémi Louf, Morgan Funtowicz, Joe
  Davison, Sam Shleifer, Patrick von Platen, Clara Ma, Yacine Jernite, Julien
  Plu, Canwen Xu, Teven~Le Scao, Sylvain Gugger, Mariama Drame, Quentin Lhoest,
  and Alexander~M. Rush.
\newblock Transformers: State-of-the-art natural language processing.
\newblock In \emph{Proceedings of the 2020 Conference on Empirical Methods in
  Natural Language Processing: System Demonstrations}, pages 38--45, Online,
  October 2020. Association for Computational Linguistics.
\newblock URL \url{https://www.aclweb.org/anthology/2020.emnlp-demos.6}.

\bibitem[Yin et~al.(2021)Yin, Mallya, Vahdat, Alvarez, Kautz, and
  Molchanov]{nvidia}
Hongxu Yin, Arun Mallya, Arash Vahdat, Jose~M. Alvarez, Jan Kautz, and Pavlo
  Molchanov.
\newblock See through gradients: Image batch recovery via gradinversion.
\newblock In \emph{CVPR}, 2021.

\bibitem[Zhao et~al.(2020)Zhao, Mopuri, and Bilen]{idlg}
Bo~Zhao, Konda~Reddy Mopuri, and Hakan Bilen.
\newblock idlg: Improved deep leakage from gradients, 2020.

\bibitem[Zhu and Blaschko(2021)]{zhu2021rgap}
Junyi Zhu and Matthew~B. Blaschko.
\newblock {R-GAP:} recursive gradient attack on privacy.
\newblock In \emph{{ICLR}}, 2021.

\bibitem[Zhu et~al.(2019)Zhu, Liu, and Han]{dlg}
Ligeng Zhu, Zhijian Liu, and Song Han.
\newblock Deep leakage from gradients.
\newblock In \emph{NeurIPS}, 2019.

\bibitem[Zhu et~al.(2015)Zhu, Kiros, Zemel, Salakhutdinov, Urtasun, Torralba,
  and Fidler]{Zhu_2015_ICCV}
Yukun Zhu, Ryan Kiros, Rich Zemel, Ruslan Salakhutdinov, Raquel Urtasun,
  Antonio Torralba, and Sanja Fidler.
\newblock Aligning books and movies: Towards story-like visual explanations by
  watching movies and reading books.
\newblock In \emph{The IEEE International Conference on Computer Vision
  (ICCV)}, December 2015.

\end{thebibliography}
\bibliographystyle{plainnat}
\section*{Checklist}
\begin{enumerate}

	\item For all authors...
	\begin{enumerate}
		\item Do the main claims made in the abstract and introduction accurately reflect the paper's contributions and scope?
		\answerYes{}
		\item Did you describe the limitations of your work?
		\answerYes{ We demonstrate the effectiveness of existing defenses against our attack in \cref{sec:eval}, further we outline that our work does not deal with reconstructing labels which is left for future work, as described in \cref{sec:method}. }
		\item Did you discuss any potential negative societal impacts of your work?
		\answerYes{ We provide a discussion in \cref{app:societal}.}
		\item Have you read the ethics review guidelines and ensured that your paper conforms to them?
		\answerYes{}
	\end{enumerate}

	\item If you are including theoretical results...
	\begin{enumerate}
		\item Did you state the full set of assumptions of all theoretical results?
		\answerNA{}
		\item Did you include complete proofs of all theoretical results?
		\answerNA{}
	\end{enumerate}

	\item If you ran experiments...
	\begin{enumerate}
		\item Did you include the code, data, and instructions needed to reproduce the main experimental results (either in the supplemental material or as a URL)?
		\answerYes{}
		\item Did you specify all the training details (e.g., data splits, hyperparameters, how they were chosen)?
		\answerYes{}
		\item Did you report error bars (e.g., with respect to the random seed after running experiments multiple times)?
		\answerYes{}
		\item Did you include the total amount of compute and the type of resources used (e.g., type of GPUs, internal cluster, or cloud provider)?
		\answerYes{We have reported the type of resources used in the \cref{sec:eval}. We have reported the total amount of compute in \cref{app:runtime}.}
	\end{enumerate}

	\item If you are using existing assets (e.g., code, data, models) or curating/releasing new assets...
	\begin{enumerate}
		\item If your work uses existing assets, did you cite the creators?
		\answerYes{}
		\item Did you mention the license of the assets?
		\answerNo{ We use standard datasets and cite the authors instead. }
		\item Did you include any new assets either in the supplemental material or as a URL?
		\answerNo{}
		\item Did you discuss whether and how consent was obtained from people whose data you're using/curating?
		\answerNA{}
		\item Did you discuss whether the data you are using/curating contains personally identifiable information or offensive content?
		\answerNo{We use standard datasets. We refer the reader to the original authors of the dataset for this discussion.}
	\end{enumerate}

	\item If you used crowdsourcing or conducted research with human subjects...
	\begin{enumerate}
		\item Did you include the full text of instructions given to participants and screenshots, if applicable?
		\answerNA{}
		\item Did you describe any potential participant risks, with links to Institutional Review Board (IRB) approvals, if applicable?
		\answerNA{}
		\item Did you include the estimated hourly wage paid to participants and the total amount spent on participant compensation?
		\answerNA{}
	\end{enumerate}

\end{enumerate}

\message{^^JLASTREFERENCESPAGE \thepage^^J}

\ifincludeappendixx
	\newpage
	\appendix
	\onecolumn
	\clearpage
\vbox{%
    \hsize\textwidth
    \linewidth\hsize

    \vskip 0.1in

    \hrule
    \vskip 0.25in
    \vskip -\parskip%

    \centering
    {\LARGE\bf Supplementary Material\par}

    \vskip 0.29in
    \vskip -\parskip
    \hrule

    \vskip 0.09in

    \vskip 0.1in
}
\section{Discussion}
\subsection{Threat Model Discussion} \label{app:threat}

In this section, we further discuss the threat model chosen by LAMP and compare it to the related work. 
To make our attack as generic as possible, we relax common assumptions that have been exploited to reconstruct client's data in the literature before. In particular, LAMP is applicable even if:
\begin{itemize}
	\item The model's word embeddings are not fine-tuned. As the gradients of the word embedding vectors are non-zero for words that are contained in the client's training data and zero otherwise, revealing the gradients to the server will allow it to easily obtain the client sequence up to reorder. This constitutes a serious breach of clients' privacy and, thus, we assume the word embeddings are not trainable.
	\item The model's positional embeddings are not fine-tuned. Similarly to the word-embedding gradients, \citet{april} have recently demonstrated that for batch size $B=1$ positional-embedding gradients can leak client's full sequence. To this end, we assume models without trainable positional embeddings.
	\item The model's transformer blocks contain no bias terms. \citet{april} have also shown that the popular attack by \citet{phong2017privacy} can be applied on the bias terms of transformer blocks to leak the client's data. To this end, we assume models without transformer block biases.
	\item The transformer model is fine-tuned on a classification task. As language modeling tasks are usually self-supervised, they often feed the same data to the model both as inputs and outputs. Based on this observation, \citet{fowl2022decepticons} have recently shown that label reconstruction algorithms can be used to obtain the client's word counts. We thus assume the more challenging binary classification setting.
	\item The server is honest-but-curious, \ie it aims to learn as much as possible about clients' data from gradients but does not tamper with the learning protocol. While prior work has shown that a malicious server can force a client to leak much more data~\cite{fowl2022decepticons}, this is orthogonal to our work. We focus on the honest-server setting instead, which is the harder setting to attack.
\end{itemize}
Note that the assumptions we make for our transformer networks can result in a small amount of accuracy loss on the final fine-tuned model, but preserve the client's data privacy much better. We emphasize that while LAMP focuses on attacks in the harder setting, it is also applicable to the simpler settings without modification.
\subsection{Improvements over Prior Work}
In this section, we outline the differences between LAMP and TAG, and we discuss how these differences help LAMP to significantly improve its text data reconstruction from gradients compared to TAG. 

A major difference between the two methods is the introduction of our discrete optimization that takes advantage of a GPT-2 language model to help reconstruct the token order better than TAG. Our discrete optimization step is novel in several ways:
\begin{itemize}
 \item It is based on a set of discrete transformations that fix common token reordering problems arising from the continuous reconstruction.
 \item We take advantage of the perplexity computed by the existing language models such as GPT-2 to evaluate the quality of different discrete transformations.
 \item Finally, our discrete optimization is alternated with the continuous, allowing for both take advantage of the result of the other which ultimately results in better token order reconstruction (See our ablation in \cref{sec:eval}).
\end{itemize}
The other major difference between our method and existing work is the choice of the error function $\mathcal{L}_\text{rec}$ used in the continuous part of the optimization. Our choice of reconstruction loss results in better reconstruction of individual tokens and thus increases R-1. In particular, we show that the cosine error function, previously applied in the image domain, can often outperform the error function suggested by TAG for text reconstruction and introduce a regularization term $\mathcal{L}_\text{reg}$ that helps the continuous optimization to converge faster to more accurate embeddings using prior knowledge about the vector sizes of embeddings. 

\section{Detailed Text Reconstruction Example} \label{app:stepbystep}

\begin{table*}[t]\centering

    \caption{Visualization of intermediate steps of text reconstruction from gradients, on a sequence from the \cola~dataset. Note that TAG performs 2500 steps, as opposed to \lampcos{} which terminates at 2000, as this is usually sufficient for convergence.} \label{table:stepbystep}
  
    \newcommand{\threecol}[1]{\multicolumn{3}{c}{#1}}
    \newcommand{\fivecol}[1]{\multicolumn{5}{c}{#1}}
    \newcommand{\ninecol}[1]{\multicolumn{9}{c}{#1}}
  
    \newcommand{\bsz}{Batch Size~}
  
    \renewcommand{\arraystretch}{1.2}
  
    \newcommand{\ccellt}[2]{\colorbox{#1}{\makebox(20,8){{#2}}}}
    \newcommand{\ccellc}[2]{\colorbox{#1}{\makebox(8,8){{#2}}}}
    \newcommand{\ccells}[2]{\colorbox{#1}{\makebox(55,8){{#2}}}}
  
    \newcommand{\noopcite}[1]{} 
    
    \resizebox{\linewidth}{!}{
      \begin{tabular}{@{}lrl@{}} \toprule
  
     Iteration & TAG & \lampcos \\ \midrule
    0 & billie icaohwatch press former spirit technical & trinity jessie maps extended evidence private peerage whatever \\  \midrule
    500 & enough stadium six too 20 le was, & many marbles have six. too. \\ \midrule
    1000 & have stadium seven too three le. marble & ; have six too many marbles. \\ \midrule
    1500 & have respect six too manys, marble & have six. too many marbles. \\ \midrule
    2000 & have... six too many i, marble & have six. too many marbles. \\ \midrule
    2500 & have... six too manys, marble & \\ \midrule
    \textcolor{refcolor}{Reference} & {\textcolor{refcolor}{i have six too many marbles.}} & {\textcolor{refcolor}{i have six too many marbles.}} \\
      \bottomrule
      \end{tabular}}
  \end{table*}

In \cref{table:stepbystep}, we show the intermediate steps of text reconstruction for a real example taken from our experiments presented in \cref{table:sentences}. We can observe that \lampcos{} reaches convergence significantly faster than the TAG baseline, and that after only 500 iterations most words are already reconstructed by our method.

\section{Additional Experiments}

\subsection{Dependency of Experimental Results to the Chosen Sentences}\label{app:senteces}
Throughout this paper, we conducted our experiments on the same $100$ sequences randomly chosen from the test portion of the datasets we attack. In this experiment, we show that our results are consistent when different sets of $100$ sequences are used. To achieve this, we ran the $\text{BERT}_\text{BASE}$ CoLA experiment with B=1 on additional 10 different sets of 100 randomly chosen sentences from the COLA test set. We report the mean $\pm$ one standard deviation of the resulting R-1, R-2 and R-L metrics averaged across the $10$ sets in \cref{table:seeds}. We see that the results are consistent with our original findings.

\begin{table}[t]\centering
	\tiny 
	
	\caption{In this experiment we reconstruct 100 random selected sentences with our methods and the baselines on the CoLA dataset and the \bertbase~($B$=1) model 10 times with 10 different randomly selected set of sentences. We report the mean and standard deviation of all ROUGE measures.} \label{table:seeds}
	
	\newcommand{\threecol}[1]{\multicolumn{3}{c}{#1}}
	\newcommand{\fivecol}[1]{\multicolumn{5}{c}{#1}}
	\newcommand{\ninecol}[1]{\multicolumn{9}{c}{#1}}
	
	\newcommand{\bsz}{Batch Size~}
	
	\renewcommand{\arraystretch}{1.2}
	
	\newcommand{\ccellt}[2]{\colorbox{#1}{\makebox(20,8){{#2}}}}
	\newcommand{\ccellc}[2]{\colorbox{#1}{\makebox(8,8){{#2}}}}
	\newcommand{\ccells}[2]{\colorbox{#1}{\makebox(55,8){{#2}}}}
	
	\newcommand{\noopcite}[1]{}
	\newcommand{\skiplen}{0.001\linewidth} 
	
	\resizebox{0.7\columnwidth}{!}{
		\begin{tabular}{@{}l rrr} \toprule
			& \makecell[c]{R-1} & \makecell[c]{R-2} & \makecell[c]{R-L} \\      
			\midrule
			\text{DLG}	& $56.2\pm5.0$ &	$6.5\pm1.6$ & $45.0\pm2.6$ \\
			\text{TAG}	& $74.4\pm3.1$ &	$10.7\pm1.8$ & $53.0\pm2.1$ \\
			\lampcos & $87.8\pm2.6$ &	$48.4\pm5.5$  & $74.6\pm2.9$ \\
			\lampell	& $83.1\pm3.7$ &	$40.7\pm5.7$ & $69.3\pm3.6$ \\
			\bottomrule
		\end{tabular}
	}
	\vspace{-0em}
\end{table}
\subsection{Attacking Gradient Masking Defense} \label{app:masking}
We experimented with a defense which zeroes out a percentage of elements in the gradient vector. In \cref{table:zeros} we vary the percentage and report MCC, R-1 and R-2. While zeroing out most gradients weakens the attack, it also reduces utility (MCC) of the model.

\begin{table}[t]
	\caption{This experiment shows the trade-off between the final network accuracy (measured by MCC) and the reconstruction quality from gradients with different percentages of zeroed-out gradient entries on the CoLA dataset on \bertbase~($B$=1).}
	\begin{center}
		\resizebox{0.45\columnwidth}{!}{
		\begin{tabular}{lrrrrrr}
			\toprule
			Zeroed \% & \makecell[c]{MCC} & \makecell[c]{R-1} & \makecell[c]{R-2} & \makecell[c]{R-L} \\
			\midrule
			$0$ & $0.557$ & $89.6$ & $51.9$ & $76.2$\\
			$75$ & $0.557$ & $79.4$ & $34.5$ & $66.3$ \\
			$90$ & $0.534$ &  $61.9$ & $20.1$ & $53.9$ \\
			$95$ & $0.515$ &  $39.0$ & $5.8$ & $37.4$ \\
			$99$ & $0.371$ & $24.7$ & $0.0$ & $24.7$ \\
			 
			\bottomrule
		\end{tabular}}
	\end{center}
	\label{table:zeros}
\end{table}
\subsection{Dependence on the Number of Initializations}\label{app:inits}
\begin{table*}[t]\centering
	
	\caption{ This experiment shows the effect on reconstruction of the chosen number of initializations $n_\text{init}$ used in \lampcos{} on all 3 datasets for \bertbase~($B$=1).} \label{table:inits}
	
	\newcommand{\threecol}[1]{\multicolumn{3}{c}{#1}}
	\newcommand{\fivecol}[1]{\multicolumn{5}{c}{#1}}
	\newcommand{\ninecol}[1]{\multicolumn{9}{c}{#1}}
	
	\newcommand{\bsz}{Batch Size~}
	\newcommand{\certified}{{CR(\%)}}
	
	\renewcommand{\arraystretch}{1.2}
	
	\newcommand{\ccellt}[2]{\colorbox{#1}{\makebox(20,8){{#2}}}}
	\newcommand{\ccellc}[2]{\colorbox{#1}{\makebox(8,8){{#2}}}}
	\newcommand{\ccells}[2]{\colorbox{#1}{\makebox(55,8){{#2}}}}
	
	\newcommand{\temp}[1]{\textcolor{red}{#1}}
	\newcommand{\noopcite}[1]{} 
	
	\newcommand{\skiplen}{0.01\linewidth} 
	\newcommand{\rlen}{0.01\linewidth} 
	
	\resizebox{0.6\linewidth}{!}{
		\begingroup
		\setlength{\tabcolsep}{4pt} %
		\begin{tabular}{@{}l p{\skiplen}  rrr p{\skiplen}  rrr p{\skiplen} rrr p{\skiplen} rrr@{}} \toprule

			&& \threecol{\cola} && \threecol{\sst} && \threecol{\rotten}\\
			
			\cmidrule(l{5pt}r{5pt}){3-5} \cmidrule(l{5pt}r{5pt}){7-9} \cmidrule(l{5pt}r{5pt}){11-13}
			
			$n_\text{init}$ && R-1 & R-2 & R-L && R-1 & R-2 & R-L && R-1 & R-2 & R-L \\ \midrule
			
			1 && 87.3 & 48.1 & 73.2 && 87.4 & {\bf 60.8} & {\bf 78.8} && 63.7 & {\bf 16.6}& {\bf 43.8}\\
			500 && {\bf 89.6}& {\bf 51.9}& {\bf 76.2} && {\bf 88.8}& 56.9 & 77.7 && {\bf 64.7} & 16.3 & 43.1\\
			\bottomrule
		\end{tabular}
		\endgroup
	} 
	
\end{table*}

In this section, we investigate the influence of our proposed initialization on the reconstructions of \lampcos{} by comparing a single random initialization ($n_\text{init}=1$) with using our two-step initialization procedure with $n_\text{init}=500$ on the \bertbase{} model and batch size of $1$. The results are shown in \cref{table:inits}. We observe that the two-step initialization scheme consistently improves individual token recovery (measured in terms of R-1) but may in some cases slightly degrade token ordering results (measured in terms of R-2). Even though we used two-step initialization in the paper (it is strictly better on one dataset and non-comparable to single random initialization on the remaining datasets), it is indeed sometimes possible to get slightly better R-2 results with the latter.

\section{Additional Experimental Details}\label{app:addexpdet}
We run all of our experiments on a single NVIDIA RTX 2080 Ti GPU with 11 GB of RAM, except for the experiments on \bertlarge{} for which we used a single NVIDIA RTX 3090 Ti GPU with 24 GB of RAM instead.

As we explain in \cref{sec:eval}, we choose the hyperparameters of our methods using a grid search approach on the \cola{} and \rotten{} datasets. For \cola{}, we first evaluated $50$ hyperparameter combinations on $10$ randomly selected (in a stratified way with respect to length) sequences from the training set (after removing the $100$ test sequences). Then, we further evaluated the best 10 combinations on different $20$ sequences from the training set. For \rotten{}, we picked the hyperparameters from the same 10 best combinations and evaluated them on the same $20$ additional sequences. For both \lampcos{} and the baselines, we investigated the following ranges for the hyperparameters: $\alpha_\text{lm} \in  [0.05, 0.2]$, $\alpha_\text{reg} \in  [0.01, 1]$, $\lambda \in [0.001, 0.5]$, $\gamma \in  [0.8, 1]$, and $\alpha_\text{tag} \in [10^{-5}, 10^{2}]$. For \lampell{}, we consider $\alpha_\text{lm} \in [30, 240]$ and $\alpha_\text{reg} \in [10, 100]$, as the scale of the loss values is orders of magnitude larger than in \lampcos{}. We experimentally found that our algorithm is robust with respect to the exact values of $n_c$ and $n_d$, provided that they are sufficiently large. To this end, we select $n_d=200$ because we found that the selected token order from the 200 random transformations is close to the optimal one according to $\mathcal{L}_{\text{rec}}(\vx) + \alpha_{\text{lm}} \mathcal{L}_{\text{lm}}(\vt)$ for the sentence lengths present in our datasets. Similarly, we set $n_c=75$ ($n_c=200$ for \bertlarge{}) which allows our continuous optimization to significantly change the embeddings before applying the next discrete optimization step in the process. Finally, we also observed the performance of our algorithm is robust with respect to $n_\text{init}$, so we set it to $500$ throughout the experiments. We note that compared to TAG and DLG, the only additional hyperparameters we have to search over are $\alpha_\text{reg}$ and $\alpha_\text{lm}$ which makes the grid search feasible for our methods.

The resulting hyperparameters for \lampcos{} are $\alpha_{\text{lm}}=0.2$, $\alpha_{\text{reg}}=1.0$, $\lambda=0.01$, $\gamma=0.89$. In contrast, the best hyperparameters for \lampell{} are $\alpha_{\text{tag}}=0.01$, $\alpha_{\text{lm}}=60$, $\alpha_{\text{reg}}=25$, $\lambda=0.01$, $\gamma=0.89$, as the loss $\mathcal{L}_{\text{tag}}$ is on a different order of magnitude compared to $\mathcal{L}_{\text{cos}}$.
In our experiments, TAG's best hyperparameters are $\alpha_{\text{tag}}=0.01$,  $\lambda=0.1$, $\gamma=1.0$ (no decay), which we also use for DLG (with $\alpha_{\text{tag}}=0.0$). 

To account for the different optimizer used in \bertlarge{} experiments, we tuned the learning rate $\lambda$ for all methods separately in this setting by evaluating each method on 5 different learning rates in the range $[0.01, 0.1]$ on 10 randomly selected sentences from the \cola{} dataset. This resulted in $\lambda=0.1$ for DLG, and $\lambda=0.01$ for TAG, \lampcos{}, and \lampell{}. We applied the chosen values of $\lambda$ to all 3 datasets. Additionally, following \citet{breaching} we clip the gradient magnitudes $\|\nabla_{\vx}  \mathcal{L}_{\text{rec}}(\vx)\|_2$ for our \bertlarge{} experiments to $1.0$ for DLG and TAG and $0.5$ for \lampcos{} and \lampell{}.
\section{Total Runtime of the Experiments}\label{app:runtime}
The \bertlarge{} model experiments in \cref{table:main} were the most computationally expensive to execute. They took between 50 hours per experiment for the \lampcos{} and \lampell{} methods and 70 hours for TAG, which executes two times more continuous optimization steps. Our experiments on the rest of the networks for both the baselines and our methods on batch size 1 ($B=1$) all took between 8 and 16 hours to execute on a single GPU with our methods being up to 2x slower due to the additional computational cost of our discrete optimization. Additionally, our experiments on batch size 4 ($B=4$) took between 8 and 36 hours to execute on a single GPU with our methods being up to 4x slower due to the additional computational cost of our discrete optimization.

\section{Potential Negative Societal Impact of This Work}\label{app:societal}
Our work is closely related to the existing works on gradients leakage attacks~(See \cref{sec:related}) which are capable of breaking the privacy promise of FL \eg \citet{idlg,geiping,nvidia,tag}. Similar to these works, \tool{} can be used to compromise the privacy of client data in real-world FL setups, especially when no defenses are used by the clients. Our attack emphasizes that text data, which is commonly used in federated settings~\cite{ramaswamy2019keyboard}, is highly vulnerable to gradient leakage attacks, similarly to data in other domains, and that when FL is applied in practice extra steps need to be taken to mitigate the potential risks. Further, in line with the related work, we study a range of possible mitigations to our attack in \cref{table:main,table:defended,table:zeros} in the paper, thus promoting possible practical FL implementations that will be less vulnerable including those defended with Gaussian noise and gradient pruning and those using bigger batch sizes.

\fi

\end{document}